\documentclass[journal,twoside,web]{ieeecolor}
\usepackage[table,xcdraw]{xcolor}

\usepackage{tmi}
\usepackage{cite}
\usepackage{amsmath,amssymb,amsfonts}
\usepackage{algorithmic}
\usepackage{graphicx}
\usepackage{textcomp}
\usepackage[hidelinks]{hyperref}
\usepackage{cleveref}
\usepackage{multirow}
\usepackage{booktabs}
\usepackage{subcaption} 
\usepackage{dsfont}

\usepackage{colortbl}

\AtBeginDocument{%
  \definecolor{lightblue}{rgb}{0.678,0.847,0.902}%
}

\usepackage{pifont}
\newcommand{\circled}[1]{\textcircled{\raisebox{-0.1em}{\small #1}}}

\newcommand{\pmstd}[2]{#1{\tiny{$\pm$#2}}}

\usepackage{siunitx}
\usepackage{makecell}
\def\BibTeX{{\rm B\kern-.05em{\sc i\kern-.025em b}\kern-.08em
    T\kern-.1667em\lower.7ex\hbox{E}\kern-.125emX}}
\begin{document}
\title{Harmonizing Generalization and Specialization: Uncertainty-Informed Collaborative Learning for Semi-supervised Medical Image Segmentation}
\author{Wenjing Lu, Yi Hong, \IEEEmembership{Member, IEEE}, Yang Yang, \IEEEmembership{Member, IEEE}
\thanks{This work was supported by the National Natural Science Foundation of China (NSFC 62272300 and 62203303). Wenjing Lu, Yi Hong, and Yang Yang are with the AGI Institute, School of Computer Science, Shanghai Jiao Tong University, Shanghai 200240, China. Yi Hong (e-mail: yi.hong@sjtu.edu.cn) and Yang Yang (e-mail: yangyang@cs.sjtu.edu.cn) are co-corresponding authors.}} 

\maketitle

\begin{abstract}
Vision foundation models have demonstrated strong generalization in medical image segmentation by leveraging large-scale, heterogeneous pretraining. However, they often struggle to generalize to specialized clinical tasks under limited annotations or rare pathological variations, due to a mismatch between general priors and task-specific requirements.
To address this, we propose Uncertainty-informed Collaborative Learning (UnCoL), a dual-teacher framework that harmonizes generalization and specialization in semi-supervised medical image segmentation.
Specifically, UnCoL distills both visual and semantic representations from a frozen foundation model to transfer general knowledge, while concurrently maintaining a progressively adapting teacher to capture fine-grained and task-specific representations. 
To balance guidance from both teachers, pseudo-label learning in UnCoL is adaptively regulated by predictive uncertainty, which selectively suppresses unreliable supervision and stabilizes learning in ambiguous regions.
Experiments on diverse 2D and 3D benchmarks show that UnCoL {consistently outperforms existing methods across most datasets and metrics, while achieving comparable performance in only a few cases.}
Moreover, our model delivers near fully supervised performance with markedly reduced annotation requirements. {Code is available at: \href{https://github.com/VivienLu/UnCoL}{https://github.com/VivienLu/UnCoL}.}
\end{abstract}


\section{Introduction}
\label{sec:introduction}
\IEEEPARstart{M}{edical} image segmentation is a cornerstone in numerous clinical workflows, including diagnosis, treatment planning, and disease monitoring~\cite{wang2022medical}. While deep learning has advanced segmentation performance, its success heavily relies on access to large-scale annotated datasets~\cite{hatamizadeh2021swin}. However, expert annotations are costly and time-consuming to obtain, especially for rare pathologies or anatomically complex structures~\cite{tarvainen2017mean,yu2019uncertainty}. This scarcity of labels motivates techniques that can learn effectively from limited supervision.

\begin{figure}
    \centering
    \includegraphics[width=\linewidth]{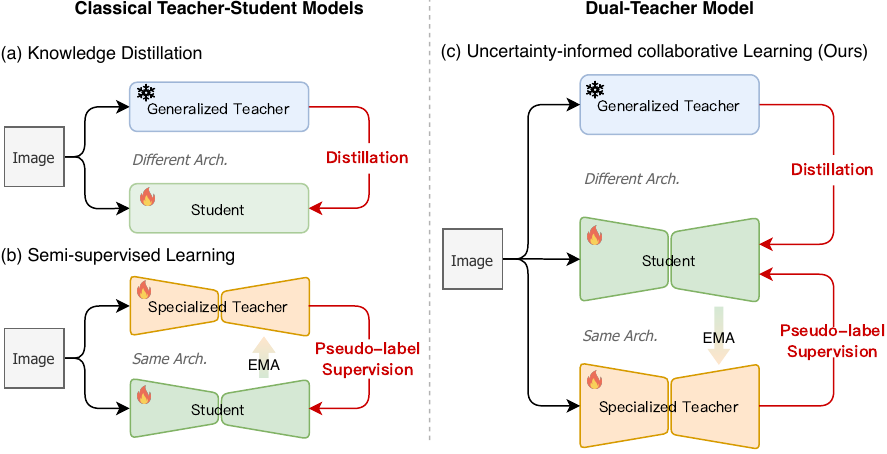}
    \caption{Comparison of teacher-student paradigms: (a) Knowledge distillation with a generalized teacher; (b) Semi-supervised learning with an EMA teacher; (c) Our dual-teacher design that integrates both models through Uncertainty-informed Collaborative Learning (UnCoL). 
    }
    \label{fig:teachers}
\end{figure}

Recent vision foundation models~\cite{kirillov2023segment,ma2024segment,wang2023sam} have demonstrated impressive generalization across organs and modalities via large-scale, heterogeneous pretraining. These models enable prompt-driven segmentation without task-specific retraining, making them attractive for low-resource scenarios~\cite{beyer2022knowledge}. Nonetheless, their effectiveness in clinical workflows remains constrained by several factors. First, their reliance on manual prompts limits applicability in fully automated pipelines. Second, their general-purpose representations often lack the task-specific precision needed for rare conditions or domain-shifted data. Although parameter-efficient tuning methods~\cite{azizi2022robust,zhang2024blo} have been explored, their performance remains sensitive to prompt quality and can degrade under sparse annotations.

In parallel, semi-supervised learning (SSL) has been extensively studied to alleviate annotation demands by leveraging unlabeled data, employing techniques such as consistency regularization~\cite{xu2023ambiguity,chi2024adaptive}, self-training~\cite{bai2023bidirectional,su2024mutual}, and uncertainty estimation~\cite{yu2019uncertainty,luo2022semi}. However, these methods are typically developed independently of foundation models, limiting their capacity to incorporate external knowledge priors. Recent SAM-enhanced SSL methods~\cite{miao2024cross,zhang2024semisam,lu2024up} attempt to bridge this gap by guiding foundation model predictions through handcrafted or model-generated prompts. However, they often assume that foundation models can accurately capture task-specific semantics. This assumption may not hold under domain shifts or complex anatomical variations, which can lead to unreliable predictions and subsequently inaccurate pseudo-labels for SSL in the absence of explicit reliability assessment.

These observations highlight a critical gap: 
\textit{foundation models offer generalizable priors but lack task-specific precision, whereas classical SSL methods focus on individual tasks and exploit unlabeled data, yet fail to fully utilize external knowledge}. 
This limitation becomes particularly evident in clinical settings characterized by limited annotations, rare pathologies, or domain shifts. 
An effective semi-supervised framework should therefore integrate the broad generalization ability of foundation models with the task-specific adaptability of conventional learning, while maintaining robustness to uncertainty under minimal supervision.

To address this, we propose \textbf{Uncertainty-Informed Collaborative Learning (UnCoL)}, a novel \textit{dual-teacher} framework for semi-supervised medical image segmentation. As illustrated in Fig.~\ref{fig:teachers}, UnCoL departs from conventional single-teacher paradigms by explicitly harmonizing generalization and specialization. 
It pairs a frozen foundation model as a \textit{generalized teacher}, providing domain-general priors, with a \textit{specialized teacher} that evolves alongside the student via exponential moving averaging (EMA) to capture task-specific knowledge. The student model learns from both teachers through two complementary mechanisms. First, a \textbf{dual-path knowledge distillation} process aligns multi-level visual and prompt-conditioned features from generalized teacher with the student’s internal representations. Second, an \textbf{uncertainty-aware pseudo-labeling} strategy adaptively selects supervision from the more reliable teacher at each unlabeled region. This collaborative scheme enables the student to integrate broad priors with fine-grained expertise, achieving robust performance under limited supervision and domain shifts. 

We highlight the main contributions of this study as follows:
\begin{itemize}
    \item[1)] We propose \textbf{UnCoL}, a semi-supervised segmentation framework that harmonizes generalization and specialization by combining the broad priors of a foundation model with the task-specific expertise of an adaptive teacher.
    \item[2)] We introduce a \textbf{dual-path knowledge distillation} strategy to transfer both visual and semantic representations to a prompt-free student, and design an \textbf{uncertainty-aware pseudo-labeling} mechanism that dynamically assigns reliable supervision to unlabeled samples.
    \item[3)] We conduct comprehensive experiments on diverse 2D and 3D medical segmentation benchmarks, demonstrating that UnCoL {achieves strong and competitive }performance under limited annotation settings.
\end{itemize}

\begin{figure*}
    \centering
    \includegraphics[width=\textwidth]{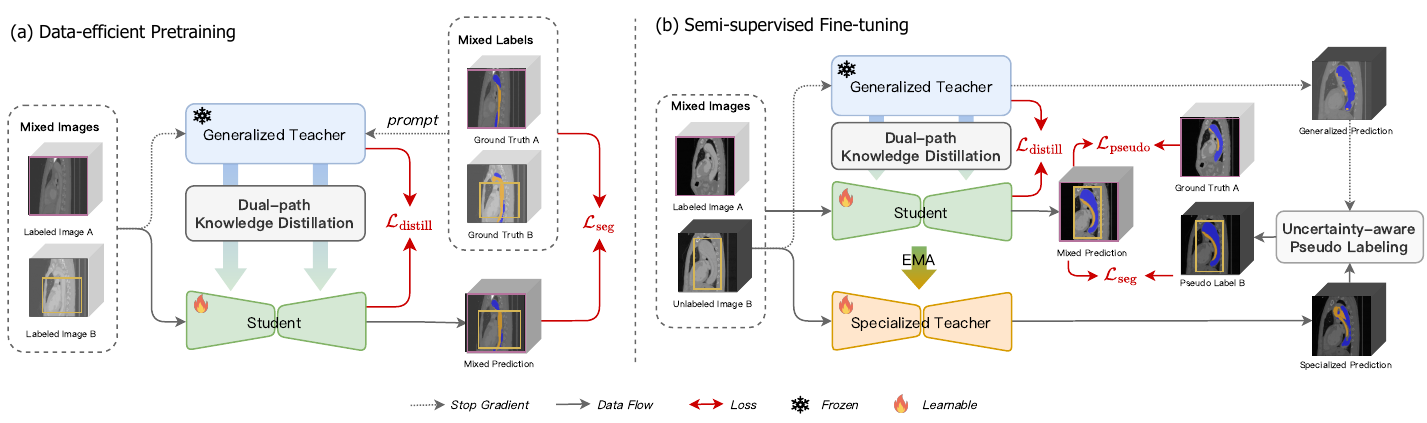}
    \caption{
    Overview of our proposed UnCoL framework. (a) In the data-efficient pretraining stage, the student learns from labeled data by distilling generalizable features from a frozen foundation model (generalized teacher) via dual-path knowledge distillation. (b) In the semi-supervised fine-tuning stage, both labeled and unlabeled data are used. Pseudo labels are assigned based on uncertainty-aware selection between the generalized teacher and the EMA-based specialized teacher. Patch-wise mixing is applied in both stages to enhance data diversity. Only the student model is used for inference as shown in Fig.~\ref{fig:infer}.
    }
    \label{fig:overview}
\end{figure*}

\section{Related Work}
\label{sec:relatedwork}
\subsection{Foundation Models in Medical Image Segmentation}
Vision foundation models such as SAM~\cite{kirillov2023segment}, MedSAM~\cite{ma2024segment}, and SAM-Med3D~\cite{wang2023sam} demonstrate promising generalization in medical image segmentation by leveraging large-scale, heterogeneous pretraining across diverse modalities and anatomical structures. Through sparse prompts (e.g., points or boxes), these models enable zero-shot or few-shot segmentation without task-specific retraining. 
To improve adaptability to downstream tasks, a variety of adaptation strategies have been explored, including full or partial fine-tuning~\cite{azizi2022robust}, architectural modifications~\cite{chen2024ma}, and parameter-efficient methods~\cite{zhang2024blo}. 
However, these methods often remain prompt-dependent and fail to deliver task-specific precision under domain shifts or limited annotations. This underscores the need for frameworks that can leverage foundation model priors more autonomously and with greater label efficiency.

\subsection{Semi-supervised Medical Image Segmentation}\label{sec:rw-ssmis}
Semi-supervised learning (SSL) is widely adopted in medical image segmentation to alleviate annotation burdens by leveraging unlabeled data alongside limited labeled samples. Representative strategies include consistency regularization~\cite{miao2023caussl,su2024mutual} and pseudo-labeling~\cite{chi2024adaptive,xiang2022fussnet}, with mean teacher framework~\cite{tarvainen2017mean} serving as a cornerstone method. Subsequent works have extended this paradigm by incorporating geometric transformations~\cite{lei2022semi}, prototype-based representations~\cite{xu2023ambiguity}{, \cite{lu2023upcol,wang2025uncertainty}}, and cross-subnet consistency~\cite{bai2023bidirectional}. 
To improve pseudo-label reliability, uncertainty estimation {has been commonly} introduced to filter low-confidence predictions, {where} Monte Carlo dropout~\cite{wu2022mutual}, model ensembles~\cite{su2024mutual}, and entropy-based filtering~\cite{yu2019uncertainty,luo2022semi} {effectively suppress noise from unlabeled supervision.
Despite their effectiveness, these approaches are predominantly domain-specific and do not explicitly leverage external foundation-model priors.}

More recently, vision foundation models have been incorporated into SSL pipelines to leverage their generalizable priors. For instance, CPC-SAM~\cite{miao2024cross} and SemiSAM~\cite{zhang2024semisam} generate prompts from auxiliary networks to guide SAM-based segmentation, while SemiSAM+\cite{zhang2025semisam+} adopts collaborative supervision between generalist and specialist models. SynFoC~\cite{ma2025steady} performs synergistic fine-tuning between foundation and conventional models but lacks explicit representation alignment, whereas H-SAM~\cite{cheng2024unleashing} eliminates prompts via hierarchical decoding but exhibits high computational overhead and limited scalability.
Although these approaches adopt dual-model strategies, they mainly enforce probability-level consistency between foundation and task-specific models, assuming comparable reliability.

In contrast, UnCoL introduces an uncertainty-informed collaboration that adaptively weights teacher reliability per pixel. Furthermore, rather than aligning probabilistic outputs,
UnCoL {transfers} visual and semantic {representations} from the frozen foundation encoder to a lightweight student. 
This design captures transferable priors while enabling prompt-free inference and encourages a natural balance between foundation-level generality and task-specific specialization.

\subsection{Knowledge Distillation}
Knowledge distillation (KD) transfers knowledge from high-capacity teacher models to smaller student networks, enhancing efficiency and generalization~\cite{hinton2015distilling}. Early methods focused on output-level alignment through softened predictions. Later approaches extend KD to intermediate representations, using feature activations~\cite{heo2019comprehensive}, or mutual information~\cite{ahn2019variational} to strengthen representational transfer.
KD has also been applied to transformer compression~\cite{zhang2023faster} and cross-architecture scenarios such as CNN-to-ViT distillation for segmentation~\cite{zhu2023good}. However, in medical image segmentation, distilling knowledge from prompt-conditioned foundation models to prompt-free student networks presents visual and semantic mismatches. These challenges highlight the need for more flexible and task-adaptive distillation strategies under limited supervision.

\section{Methodology} 
\label{sec:method}
\subsection{Problem Formulation}
\label{sec:method-over}
We consider a semi-supervised medical image segmentation task with $C$ semantic classes. Let $\mathcal{D}_L = \{(x_i, y_i)\}_{i=1}^{N_L}$ denote the labeled training set {of size $N_L$}, where $x_i \in \mathcal{X}$ is a 2D or 3D medical image and $y_i \in \mathcal{Y}$ is the corresponding pixel/voxel-wise segmentation mask. Let $\mathcal{D}_U = \{x_j\}_{j=1}^{N_U}$ be a much larger set of unlabeled images, with the data size $N_U \gg N_L$. We aim to learn a segmentation model $f_\theta: \mathcal{X} \rightarrow \mathcal{Y}$, parameterized by $\theta$, that predicts masks $\hat{y} = f_\theta(x)$ by effectively leveraging both $\mathcal{D}_L$ and $\mathcal{D}_U$ under limited annotations. 
{To incorporate the foundation model into this framework, we denote it by $f_{\xi}$ and express its prompt-conditioned prediction as $\hat{y}_{\xi} = f_{\xi}(x, \pi)$, where $\pi$ is the prompt embedding. The student model $f_{\theta}(x)$ performs prompt-free inference and learns from $f_{\xi}$ through uncertainty-informed supervision, allowing transferable visual and semantic priors to be incorporated into its representation space. We then optimize $\theta$ by combining this teacher-guided supervision with the labeled segmentation loss on $\mathcal{D}_L$ and unsupervised consistency regularization terms defined on $\mathcal{D}_U$.}

\subsection{Framework Overview}
\subsubsection{Model Architecture}
As shown in Fig.~\ref{fig:overview}, UnCoL comprises three key components:

\begin{itemize}
    \item \textbf{Generalized teacher:} A frozen foundation model {(i.e., MedSAM~\cite{ma2024segment} for 2D tasks or SAM-Med3D~\cite{wang2023sam} for 3D tasks in this work)} pretrained on large-scale and diverse medical datasets. It provides stable and high-capacity guidance via visual and semantic representations.
    \item \textbf{Specialized teacher:} A task-adaptive teacher model implemented as an exponential moving average (EMA) of the student model. It captures domain-specific representations and offers temporally smoothed predictions tailored to the target distribution.
    \item \textbf{Student model:} A lightweight segmentation network with a compact transformer-based image encoder and a CNN-based segmentation backbone. It learns from both teachers via dual-path knowledge distillation and uncertainty-aware pseudo-labeling.
\end{itemize}

\subsubsection{Supervision Strategy}
UnCoL integrates dual-teacher guidance through two complementary mechanisms. The \textbf{Dual-Path Knowledge Distillation (DPKD)} module transfers generalizable priors by aligning the student encoder with visual and semantic representations from the frozen foundation model, enabling prompt-free distillation. The \textbf{Uncertainty-Aware Pseudo-Labeling (UAPL)} module selects the teacher with lower predictive uncertainty at each spatial location, enabling reliable supervision by leveraging the complementary strengths of generalization and specialization.

\subsection{Dual-Path Knowledge Distillation}
\label{sec:method-dpd}
\begin{figure}
    \centering
    \includegraphics[width=\linewidth]{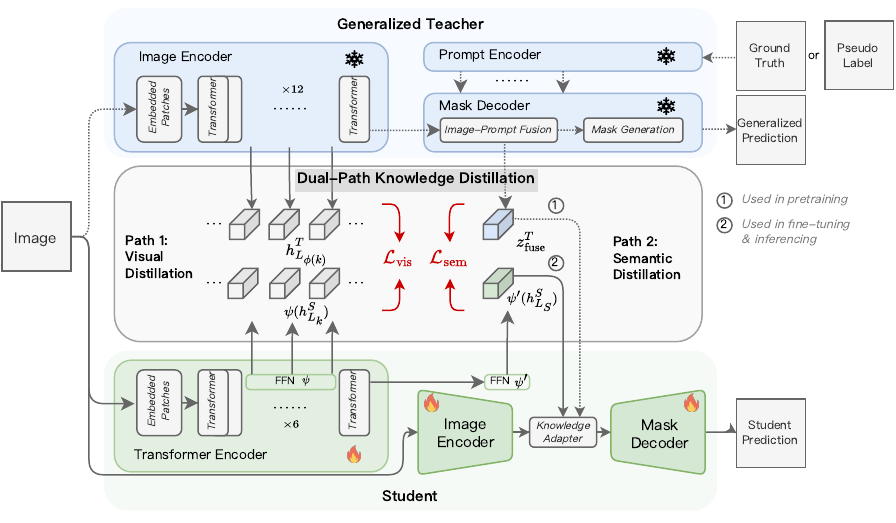}
    \caption{The Dual-path Knowledge Distillation module: Path 1 distills multi-layer visual features from the generalized teacher’s encoder to the student's encoder; Path 2 transfers prompt-conditioned semantics to the student's final layer. 
    {Here, $\phi$ is the layer-mapping function, and $\psi,\psi^\prime$ are FFN-based projection layers for dimension matching.}
    Dashed arrows (\circled{1}) indicate pretraining-only connections, and solid arrows (\circled{2}) denote fine-tuning and inference paths.
    }
    \label{fig:DPD}
\end{figure}

We introduce a Dual-Path Knowledge Distillation (DPKD) module to transfer {multi-level visual representations and prompt-informed semantics from the frozen foundation model to the student}.
DPKD {includes} two paths: (1) visual distillation from intermediate transformer layers, and (2) semantic distillation via prompt-conditioned features derived from the teacher’s image-prompt feature fusion module{ used only during pretraining}. This design enables the student to inherit visual priors and prompt-informed semantics without requiring prompts during inference. {DPKD architecture} is illustrated in Fig.~\ref{fig:DPD}. Let $h^T$ and $h^S$ denote the encoders of the teacher and student models, consisting of $L_T$ and $L_S$ layers, respectively. A set of $K$ layer pairs is selected for alignment, where a mapping function $\phi:\{1,\dots,K\}\to\{1,\dots,L_T\}$ assigns each selected student layer $L_k$ to a corresponding teacher layer $L_{\phi(k)}$. {This multi-layer mapping allows the lightweight student to inherit multi-level representations from the deeper teacher, ensuring consistency {despite differences in network depth}.}

\subsubsection{Visual Distillation Path}
The student aligns its intermediate representations with the {generalized} teacher’s by minimizing the discrepancy between corresponding feature maps. Let $h_{L_k}^S \in \mathbb{R}^{N \times D_S}$ and $h_{L_{\phi(k)}}^T \in \mathbb{R}^{N \times D_T}$ be the feature outputs of the $k$-th selected student layer and its teacher counterpart{, respectively. Here, $N$  is the number of visual tokens, which is identical for the teacher and student because both use matched patch embeddings}. If $D_S \neq D_T$, a projection layer $\psi: \mathbb{R}^{D_S} \rightarrow \mathbb{R}^{D_T}$ is applied {to project the student’s feature dimension to that of the teacher (implemented with a lightweight FFN)}. The visual distillation loss is:
\begin{equation}
\mathcal{L}_{\text{vis}} = \frac{1}{K} \sum_{k=1}^{K} \left\| \psi\left(h_{L_k}^S\right) - h_{L_{\phi(k)}}^T \right\|_2^2,
\end{equation}
which encourages the student to mimic the teacher’s hierarchical visual features.

\subsubsection{Semantic Distillation Path}
To transfer prompt-informed semantics, we extract the fused image-prompt representation $z^T_{\text{fuse}} \in \mathbb{R}^{N \times D_T^{\prime}}$ from the generalized teacher and align it with the student’s output of the last transformer encoder layer, i.e., $h_{L_S}^S \in \mathbb{R}^{N \times D_S^{\prime}}$. If $D_S^{\prime} \neq D_T^{\prime}$, {a similar projection layer $\psi^{\prime}: \mathbb{R}^{D_S^{\prime}} \rightarrow \mathbb{R}^{D_T^{\prime}}$ is employed in the semantic path to align the student’s final-layer representation with the teacher’s prompt-conditioned semantics}. Then, the semantic distillation loss is:
\begin{equation}
\mathcal{L}_{\text{sem}} = \left\| \psi^{\prime}\left(h_{L_S}^S\right) - z^T_{\text{fuse}} \right\|_2^2,
\end{equation}
which promotes alignment with the global semantic priors encoded by the prompt-informed teacher. {Unlike $\mathcal{L}_{\text{vis}}$, which aggregates losses across multiple layers, $\mathcal{L}_{\text{sem}}$ focuses solely on the final encoder layer to capture high-level prompt-informed semantics.}

The total distillation loss is given by $\mathcal{L}_{\text{distill}} = \mathcal{L}_{\text{vis}} + \mathcal{L}_{\text{sem}}.$
By jointly minimizing these objectives, the student learns to replicate both the visual structure and semantic context of the foundation model, enhancing {the student’s representation quality while maintaining a compact and prompt-free architecture}.

\subsection{Uncertainty-Aware Pseudo-Labeling}
\label{sec:method-upl}
To enable effective supervision on unlabeled data, we introduce an Uncertainty-Aware Pseudo-Labeling (UAPL) strategy that dynamically selects the more reliable teacher at each spatial location based on predictive uncertainty. This design exploits the complementary strengths of the generalized teacher for robust prior knowledge and the specialized teacher for fine-grained contextual understanding.

Given an unlabeled image $x \in \mathcal{D}_U$ defined on a spatial domain $\Omega$, the generalized and specialized teachers independently produce class probability maps $p^G(x), p^S(x) \in [0,1]^{C \times |\Omega|}$, where $C$ denotes the number of semantic classes. 
{Predictive uncertainty at each location $i \in \Omega$ is computed using Shannon entropy.}
Specifically, the uncertainties from the generalized {and specialized teachers are given by
$u^G(i) = -\sum_{c=1}^C p_c^G(i) \log p_c^G(i)$ and
$u^S(i) = -\sum_{c=1}^C p_c^S(i) \log p_c^S(i)$, respectively.}
A binary confidence mask is constructed for each teacher by thresholding entropy with a ramp-up threshold $\tau$, yielding $m^G(i) = \mathbb{I}\{ u^G(i) \leq \tau \}$ and $m^S(i) = \mathbb{I}\{ u^S(i) \leq \tau \}$, where $\mathbb{I}\{\cdot\}$ is the indicator function and $\tau$ increases over training (see Sec.~\ref{sec:exp_set} for details).

To supervise the student model, we generate a fused pseudo-probability map $\tilde{p}(i) \in [0,1]^C$ based on teachers' confidence. If both teachers are confident at location $i$, i.e., $m^G(i) = 1$ and $m^S(i) = 1$, their predictions are combined via an uncertainty-weighted average:
\begin{equation}
\tilde{p}(i) = \frac{e^{-u^G(i)} p^G(i) + e^{-u^S(i)} p^S(i)}{e^{-u^G(i)} + e^{-u^S(i)}}.
\end{equation}
If only one teacher is confident, its prediction is used directly; that is, $\tilde{p}(i) = p^G(i)$ if $m^G(i) = 1$ and $m^S(i) = 0$, or $\tilde{p}(i) = p^S(i)$ if $m^S(i) = 1$ and $m^G(i) = 0$. Locations where neither teacher is confident are {omitted} from training supervision.

Finally, hard pseudo-labels are obtained by selecting the most probable class at each valid location: $\tilde{y}(i) = \arg\max_c \tilde{p}_c(i)$. The supervised region is defined as $\Omega^* = \{ i \in \Omega \mid m^G(i) + m^S(i) \geq 1 \}$, ensuring that only pixels with reliable predictions from at least one teacher contribute to the student’s learning.
The student is optimized on these reliable regions using a hybrid segmentation loss:
\begin{equation}
\mathcal{L}_{\text{pseudo}} = \frac{1}{|\Omega^*|} \sum_{i \in \Omega^*} \left( \text{CE}(\hat{y}(i), \tilde{y}(i)) + \text{Dice}(\hat{y}(i), \tilde{y}(i)) \right).
\end{equation}

{This strategy derives pseudo-labels from confident teacher predictions, reducing noise and improving supervision quality in regions prone to uncertainty or teacher disagreement.}
\subsection{Training and Inference Strategy} 
{As illustrated in Fig.~\ref{fig:overview}, }we adopt a two-stage training scheme that progressively guides the student from domain-general priors to task-specific knowledge, followed by an efficient inference phase.

\subsubsection{Stage 1: Data-Efficient Pretraining} \label{sec:s1_pretraining}

In the first stage, the student is trained on the labeled dataset $\mathcal{D}_L$ under supervision from both ground-truth annotations and the frozen foundation model. To enhance data diversity, we apply spatial copy-paste augmentation~\cite{bai2023bidirectional}, which mixes two labeled images by replacing 60\% of spatial patches from one image with those from another. The pretraining objective combines a supervised segmentation loss $\mathcal{L}_{\text{sup}}$ and a dual-path distillation loss:
\begin{equation}
\mathcal{L}^{\text{pre}} = \mathcal{L}_{\text{sup}} + \alpha(t) \mathcal{L}_{\text{distill}}^{\text{pre}},
\end{equation}
where $\mathcal{L}_{\text{distill}}^{\text{pre}} = \mathcal{L}_{\text{vis}} + \mathcal{L}_{\text{sem}}$ is defined in Sec.~\ref{sec:method-dpd}, and $\alpha(t)$ is a time-dependent weighting schedule. 

During this stage, a knowledge adapter is applied to fuse the student’s image features with prompt-informed representations $z^T_{\text{fuse}}$ from the foundation model. The adapter is implemented as a cross-attention module, where student’s image features act as the query and teacher-provided features as key and value. This design enables semantic conditioning while preserving prompt-free inference.

\subsubsection{Stage 2: Semi-Supervised Fine-Tuning}

In the second stage, both labeled and unlabeled data ($\mathcal{D}_L \cup \mathcal{D}_U$) are used. To exploit spatial structures across domains, we adopt a copy-paste strategy that mixes patches from labeled images with ground-truth annotations and unlabeled images with pseudo-labels, ensuring that the label maps are mixed consistently with the image content. A specialized teacher is constructed by applying EMA to the student model’s parameters. During both training stages, its parameters are updated by $\theta_S \leftarrow \mu \theta_S + (1 - \mu)\theta$, where $\theta_S$ and $\theta$ denote the parameters of the specialized teacher and student models, respectively. The momentum coefficient is set to $\mu=0.99$, following~\cite{tarvainen2017mean}. The student continues to be supervised by $\mathcal{L}_{\text{sup}}$ on labeled data, in line with~\cite{xiang2022fussnet}, and is supervised on unlabeled samples using pseudo-labels generated via the uncertainty-aware strategy.
To preserve generalization, the visual distillation loss $\mathcal{L}_{\text{vis}}$ is reserved, while semantic distillation $\mathcal{L}_{\text{sem}}$ is omitted to avoid potential noise from synthetic prompts.

The knowledge adapter remains active during this stage. For labeled data, it incorporates prompt-informed features from the generalized teacher; for unlabeled data, it attends to high-level representations $h_{L_S}^S$ from the specialized teacher. The final objective is:
\begin{equation}
\mathcal{L}^{\text{fine}} = \mathcal{L}_{\text{sup}} + \lambda_{\text{pseudo}} \mathcal{L}_{\text{pseudo}} + \lambda_{\text{vis}} \mathcal{L}_{\text{vis}},
\end{equation}
where $\lambda_{\text{pseudo}}$ and $\lambda_{\text{vis}}$ are trade-off hyperparameters.

\subsubsection{Inference {Phase}}

\begin{figure}
    \centering
    \includegraphics[width=\linewidth]{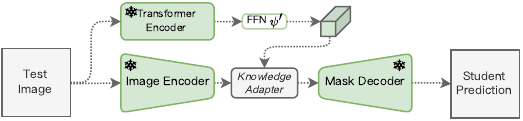}
    \caption{Inference pipeline of the student model. 
    }
    \label{fig:infer}
\end{figure}

At inference, only the student model is used. The knowledge adapter operates solely on the student’s transformer and image encoder features, using no external prompts or teacher guidance. As illustrated in Fig.~\ref{fig:infer}, segmentation is performed in a single forward pass, yielding an efficient model with strong generalization and task adaptation.

\section{Experiments}
\label{sec:exp}
\subsection{Experimental Settings} \label{sec:exp_set}

\begin{table*}[htbp]
\centering
\caption{Comparison on the OASIS dataset under varying supervision ratios. 
{``Labeled Ratio" indicates the ratio of labeled data.}
95HD and ASD are reported in voxel units.
\textbf{Bold} and \underline{underline} indicate the best and second-best results.
} \label{tab:oasis}
\resizebox{\textwidth}{!}{
\begin{tabular}{l|c|llll|llll|llll|llll|llll}
\toprule[1.3pt]
\multirow{2}{*}{Method} & \multicolumn{1}{c|}{\multirow{2}{*}{\makecell{{Labeled}\\{Ratio}}}} & \multicolumn{4}{c|}{Left Thalamus (LT)}   & \multicolumn{4}{c|}{Left Hippocampus (LH)} & \multicolumn{4}{c|}{Right Thalamus (RT)}  & \multicolumn{4}{c|}{Right Hippocampus (RH)} & \multicolumn{4}{c}{Average}  \\ \cmidrule{3-22}
 & & \multicolumn{1}{c}{\scriptsize{DSC\tiny{$(\%)$}\scriptsize{$\uparrow$}}} & \multicolumn{1}{c}{\scriptsize{Jaccard\tiny{$(\%)$}\scriptsize{$\uparrow$}}} & \multicolumn{1}{c}{\scriptsize{95HD$\downarrow$}} & \multicolumn{1}{c|}{\scriptsize{ASD$\downarrow$}} & \multicolumn{1}{c}{\scriptsize{DSC\tiny{$(\%)$}\scriptsize{$\uparrow$}}} & \multicolumn{1}{c}{\scriptsize{Jaccard\tiny{$(\%)$}\scriptsize{$\uparrow$}}} & \multicolumn{1}{c}{\scriptsize{95HD$\downarrow$}} & \multicolumn{1}{c|}{\scriptsize{ASD$\downarrow$}} & \multicolumn{1}{c}{\scriptsize{DSC\tiny{$(\%)$}\scriptsize{$\uparrow$}}} & \multicolumn{1}{c}{\scriptsize{Jaccard\tiny{$(\%)$}\scriptsize{$\uparrow$}}} & \multicolumn{1}{c}{\scriptsize{95HD$\downarrow$}} & \multicolumn{1}{c|}{\scriptsize{ASD$\downarrow$}} & \multicolumn{1}{c}{\scriptsize{DSC\tiny{$(\%)$}\scriptsize{$\uparrow$}}} & \multicolumn{1}{c}{\scriptsize{Jaccard\tiny{$(\%)$}\scriptsize{$\uparrow$}}} & \multicolumn{1}{c}{\scriptsize{95HD$\downarrow$}} & \multicolumn{1}{c|}{\scriptsize{ASD$\downarrow$}} & \multicolumn{1}{c}{\scriptsize{DSC\tiny{$(\%)$}\scriptsize{$\uparrow$}}} & \multicolumn{1}{c}{\scriptsize{Jaccard\tiny{$(\%)$}\scriptsize{$\uparrow$}}} & \multicolumn{1}{c}{\scriptsize{95HD$\downarrow$}} & \multicolumn{1}{c}{\scriptsize{ASD$\downarrow$}} \\ \midrule\midrule
SAM\tiny{(ICCV'23)~\cite{kirillov2023segment}}   & \multirow{2}{*}{\makecell{{0/289}\\{($0\%$)}}} & 39.60\tiny{$\pm$4.47} & 24.78\tiny{$\pm$3.41} & 77.82\tiny{$\pm$9.59} & 9.79\tiny{$\pm$0.53} & 20.84\tiny{$\pm$2.91} & 11.66\tiny{$\pm$1.81} & 95.54\tiny{$\pm$1.10} & 11.03\tiny{$\pm$0.73} & 51.91\tiny{$\pm$2.53} & 35.09\tiny{$\pm$2.26} & 15.03\tiny{$\pm$2.70} & 8.33\tiny{$\pm$0.55} & 39.95\tiny{$\pm$3.73} & 25.03\tiny{$\pm$2.85} & 15.16\tiny{$\pm$1.28} & 10.00\tiny{$\pm$0.67} & 38.07\tiny{$\pm$11.66} & 24.14\tiny{$\pm$8.73} & 50.89\tiny{$\pm$36.69} & 9.79\tiny{$\pm$1.15}  \\
MedSAM\tiny{(NC'24)~\cite{ma2024segment}}  & & 78.41\tiny{$\pm$7.30} & 65.09\tiny{$\pm$9.97} & 7.51\tiny{$\pm$2.18} & 2.06\tiny{$\pm$0.88} & 75.57\tiny{$\pm$8.90} & 61.54\tiny{$\pm$11.00} & 6.85\tiny{$\pm$1.99} & 2.22\tiny{$\pm$0.80} & 75.09\tiny{$\pm$9.10} & 60.82\tiny{$\pm$9.65} & 8.13\tiny{$\pm$2.10} & 2.49\tiny{$\pm$0.93} & 76.88\tiny{$\pm$8.97} & 63.28\tiny{$\pm$11.27} & 6.71\tiny{$\pm$1.87} & 2.13\tiny{$\pm$0.79} & 76.49\tiny{$\pm$8.78} & 62.68\tiny{$\pm$10.71} & 7.30\tiny{$\pm$2.13} & 2.23\tiny{$\pm$0.87} \\ \midrule\midrule
UNet\tiny{(MICCAI'15)~\cite{ronneberger2015u}}  & \multirow{9}{*}{\makecell{{15/289}\\{($\sim5\%$)}}}  & 83.54\tiny{$\pm$9.77} & 75.24\tiny{$\pm$11.65} & 14.10\tiny{$\pm$10.55} & 4.53\tiny{$\pm$3.18} & 47.65\tiny{$\pm$24.90} & 35.18\tiny{$\pm$21.50} & 64.69\tiny{$\pm$23.62} & 29.67\tiny{$\pm$15.97} & 77.89\tiny{$\pm$12.76} & 70.68\tiny{$\pm$13.85} & 8.00\tiny{$\pm$6.29} & 3.32\tiny{$\pm$2.77} & 16.34\tiny{$\pm$22.34} & 12.06\tiny{$\pm$17.96} & 17.22\tiny{$\pm$23.84} & 7.63\tiny{$\pm$14.13} & 56.36\tiny{$\pm$36.25} & 48.29\tiny{$\pm$34.05} & 26.00\tiny{$\pm$30.68} & 11.29\tiny{$\pm$16.34} \\
{nnU-Net\tiny{(Nat. Methods'21)~\cite{isensee2021nnu}}} &  & {93.35\tiny{$\pm$ 2.54}} & {87.64\tiny{$\pm$ 4.27}} & {5.93\tiny{$\pm$ 10.94}} & {2.10\tiny{$\pm$ 3.04}} & {87.11\tiny{$\pm$ 7.80}} & {77.96\tiny{$\pm$ 10.86}} & {14.08\tiny{$\pm$ 26.28}} & {4.11\tiny{$\pm$ 6.96}} & {93.27\tiny{$\pm$ 2.70}} & {87.52\tiny{$\pm$ 4.50}} & {5.45\tiny{$\pm$ 9.95}} & {1.79\tiny{$\pm$ 2.61}} & {86.63\tiny{$\pm$ 7.37}} & {77.19\tiny{$\pm$ 9.97}} & {41.17\tiny{$\pm$ 38.65}} & {13.19\tiny{$\pm$ 12.28}} & {90.09\tiny{$\pm$ 6.59}} & {82.58\tiny{$\pm$ 9.49}} & {16.66\tiny{$\pm$ 29.08}} & {5.30\tiny{$\pm$ 8.77}} \\ \cmidrule(lr){1-1} \cmidrule(lr){3-22}
MT\tiny{(NeurIPS'17)~\cite{tarvainen2017mean}} & & 81.10\tiny{$\pm$19.39} & 72.80\tiny{$\pm$21.59} & 20.84\tiny{$\pm$20.87} & 7.17\tiny{$\pm$7.38} & 71.77\tiny{$\pm$20.96} & 60.20\tiny{$\pm$22.34} & 43.21\tiny{$\pm$30.77} & 17.24\tiny{$\pm$13.07} & 80.46\tiny{$\pm$22.78} & 72.93\tiny{$\pm$23.88} & 15.50\tiny{$\pm$17.30} & 4.80\tiny{$\pm$5.57} & 57.12\tiny{$\pm$31.59} & 49.43\tiny{$\pm$29.85} & 19.71\tiny{$\pm$27.91} & 7.04\tiny{$\pm$13.70} & 72.61\tiny{$\pm$27.18} & 63.84\tiny{$\pm$27.45} & 24.82\tiny{$\pm$29.77} & 9.06\tiny{$\pm$12.62}  \\
UAMT\tiny{(MICCAI'19)~\cite{yu2019uncertainty}}  & & 83.49\tiny{$\pm$12.92} & 74.00\tiny{$\pm$15.12} & 19.20\tiny{$\pm$14.85} & 6.32\tiny{$\pm$4.92} & 67.01\tiny{$\pm$22.58} & 55.24\tiny{$\pm$23.07} & 39.68\tiny{$\pm$30.69} & 15.76\tiny{$\pm$13.17} & 79.19\tiny{$\pm$19.36} & 70.64\tiny{$\pm$20.28} & 12.85\tiny{$\pm$13.38} & 3.77\tiny{$\pm$4.37} & 55.38\tiny{$\pm$30.49} & 46.05\tiny{$\pm$28.71} & 28.45\tiny{$\pm$29.15} & 11.03\tiny{$\pm$14.42} & 71.27\tiny{$\pm$26.61} & 61.48\tiny{$\pm$26.50} & 25.05\tiny{$\pm$28.65} & 9.22\tiny{$\pm$12.50} \\
URPC\tiny{(MIA'22)~\cite{luo2022semi}}  & & 70.40\tiny{$\pm$25.24} & 62.93\tiny{$\pm$25.69} & 11.15\tiny{$\pm$12.15} & 4.35\tiny{$\pm$4.77} & 61.92\tiny{$\pm$26.94} & 51.48\tiny{$\pm$26.28} & 37.44\tiny{$\pm$28.94} & 15.92\tiny{$\pm$15.04} & 66.07\tiny{$\pm$27.17} & 59.46\tiny{$\pm$26.88} & 10.08\tiny{$\pm$10.26} & 4.13\tiny{$\pm$5.63} & 54.23\tiny{$\pm$30.75} & 45.54\tiny{$\pm$28.94} & 26.49\tiny{$\pm$30.73} & 12.15\tiny{$\pm$16.28} & 63.15\tiny{$\pm$30.93} & 54.86\tiny{$\pm$30.16} & 21.29\tiny{$\pm$27.53} & 9.14\tiny{$\pm$13.63} \\
BCP\tiny{(CVPR'23)~\cite{bai2023bidirectional}}   & & 92.06\tiny{$\pm$4.98} & 85.80\tiny{$\pm$6.97} & 3.47\tiny{$\pm$6.31} & 1.17\tiny{$\pm$1.40} & 85.49\tiny{$\pm$8.71} & 75.82\tiny{$\pm$11.26} & 7.14\tiny{$\pm$10.98} & 2.39\tiny{$\pm$3.93} & 91.13\tiny{$\pm$5.89} & 84.28\tiny{$\pm$8.12} & 3.23\tiny{$\pm$4.83} & 1.20\tiny{$\pm$1.47} & 88.24\tiny{$\pm$10.00} & 80.39\tiny{$\pm$11.54} & 7.89\tiny{$\pm$12.30} & 2.63\tiny{$\pm$4.70} & 89.23\tiny{$\pm$8.93} & 81.57\tiny{$\pm$11.28} & 5.43\tiny{$\pm$10.94} & 1.85\tiny{$\pm$3.74} \\
ABD\tiny{(CVPR'24)~\cite{chi2024adaptive}}   & & \underline{94.18\tiny{$\pm$3.14}} & \underline{89.30\tiny{$\pm$3.30}} & 2.86\tiny{$\pm$3.60} & 0.94\tiny{$\pm$1.11} & \underline{89.49\tiny{$\pm$1.34}} & \underline{81.35\tiny{$\pm$1.38}} & 5.70\tiny{$\pm$8.23} & 1.79\tiny{$\pm$1.92} & \underline{94.38\tiny{$\pm$2.84}} & \underline{89.67\tiny{$\pm$3.17}} & 3.12\tiny{$\pm$3.53} & 0.99\tiny{$\pm$0.99} & \underline{90.01\tiny{$\pm$0.80}} & \underline{82.14\tiny{$\pm$1.08}} & 5.00\tiny{$\pm$5.83} & 1.54\tiny{$\pm$1.91} & \underline{92.01\tiny{$\pm$1.56}} & \underline{85.61\tiny{$\pm$1.62}} & 4.17\tiny{$\pm$4.72} & 1.31\tiny{$\pm$1.10} \\ \cmidrule(lr){1-1} \cmidrule(lr){3-22}
CPC-SAM\tiny{(MICCAI'24)~\cite{miao2024cross}} & & 93.83\tiny{$\pm$1.57} & 88.42\tiny{$\pm$2.76} &	\underline{2.13\tiny{$\pm$0.51}} &	\underline{0.79\tiny{$\pm$0.25}} & 88.39\tiny{$\pm$2.64} & 79.30\tiny{$\pm$4.21} & \underline{2.89\tiny{$\pm$0.95}} & \underline{1.01\tiny{$\pm$0.36}} & 93.87\tiny{$\pm$1.62} & 88.49\tiny{$\pm$2.87} & \underline{2.25\tiny{$\pm$1.34}} & \underline{0.85\tiny{$\pm$0.42}} & 89.22\tiny{$\pm$3.70} & 80.73\tiny{$\pm$5.81} & \underline{2.62\tiny{$\pm$1.02}} & \underline{0.90\tiny{$\pm$0.26}} & 91.33\tiny{$\pm$3.59} & 84.24\tiny{$\pm$5.91} & \underline{2.47\tiny{$\pm$1.04}} & \underline{0.89\tiny{$\pm$0.34}} \\
\cellcolor{lightblue!50}\textbf{UnCoL(ours)}   & & \cellcolor{lightblue!50}\textbf{95.54\tiny{$\pm$1.31}} & \cellcolor{lightblue!50}\textbf{91.50\tiny{$\pm$2.31}} & \cellcolor{lightblue!50}\textbf{1.52\tiny{$\pm$1.93}} & \cellcolor{lightblue!50}\textbf{0.54\tiny{$\pm$0.39}} & \cellcolor{lightblue!50}\textbf{92.36\tiny{$\pm$3.28}} & \cellcolor{lightblue!50}\textbf{85.97\tiny{$\pm$5.19}} & \cellcolor{lightblue!50}\textbf{2.90\tiny{$\pm$4.06}} & \cellcolor{lightblue!50}\textbf{0.81\tiny{$\pm$1.22}} & \cellcolor{lightblue!50}\textbf{95.46\tiny{$\pm$1.36}} & \cellcolor{lightblue!50}\textbf{91.35\tiny{$\pm$2.42}} & \cellcolor{lightblue!50}\textbf{1.47\tiny{$\pm$0.55}} & \cellcolor{lightblue!50}\textbf{0.54\tiny{$\pm$0.15}} & \cellcolor{lightblue!50}\textbf{92.90\tiny{$\pm$3.18}} & \cellcolor{lightblue!50}\textbf{86.92\tiny{$\pm$5.05}} & \cellcolor{lightblue!50}\textbf{2.82\tiny{$\pm$5.06}} & \cellcolor{lightblue!50}\textbf{0.78\tiny{$\pm$1.25}} & \cellcolor{lightblue!50}\textbf{94.07\tiny{$\pm$2.92}} & \cellcolor{lightblue!50}\textbf{88.93\tiny{$\pm$4.78}} & \cellcolor{lightblue!50}\textbf{2.18\tiny{$\pm$4.63}} & \cellcolor{lightblue!50}\textbf{0.67\tiny{$\pm$1.28}}  \\ \midrule\midrule
UNet\tiny{(MICCAI'15)~\cite{ronneberger2015u}}  & \multirow{9}{*}{\makecell{{29/289}\\{($\sim10\%$)}}}  & 86.33\tiny{$\pm$8.15} & 77.25\tiny{$\pm$10.37} & 10.02\tiny{$\pm$9.08} & 3.00\tiny{$\pm$2.44} & 48.96\tiny{$\pm$27.54} & 38.10\tiny{$\pm$24.79} & 53.95\tiny{$\pm$28.02} & 25.57\tiny{$\pm$17.31} & 86.52\tiny{$\pm$7.24} & 77.43\tiny{$\pm$9.72} & 10.39\tiny{$\pm$7.79} & 3.31\tiny{$\pm$2.13} & 43.74\tiny{$\pm$28.93} & 34.07\tiny{$\pm$25.47} & 21.66\tiny{$\pm$20.83} & 7.47\tiny{$\pm$9.57} & 66.39\tiny{$\pm$30.16} & 56.71\tiny{$\pm$29.29} & 24.01\tiny{$\pm$28.74} & 9.84\tiny{$\pm$15.09}  \\
{nnU-Net\tiny{(Nat. Methods'21)~\cite{isensee2021nnu}}} &  & {92.71\tiny{$\pm$ 4.65}} & {86.84\tiny{$\pm$ 7.08}} & {6.57\tiny{$\pm$ 10.65}} & {2.11\tiny{$\pm$ 3.01}} & {88.84\tiny{$\pm$ 7.25}} & {80.63\tiny{$\pm$ 9.44}} & {7.59\tiny{$\pm$ 15.96}} & {2.31\tiny{$\pm$ 4.49}} & {92.03\tiny{$\pm$ 6.23}} & {86.06\tiny{$\pm$ 8.59}} & {3.91\tiny{$\pm$ 5.76}} & {1.37\tiny{$\pm$ 1.53}} & {88.91\tiny{$\pm$ 5.90}} & {80.51\tiny{$\pm$ 8.73}} & {31.18\tiny{$\pm$ 34.45}} & {9.45\tiny{$\pm$ 10.13}} & {90.62\tiny{$\pm$ 6.69}} & {83.51\tiny{$\pm$ 9.43}} & {12.31\tiny{$\pm$ 22.94}} & {3.81\tiny{$\pm$ 6.68}} \\ \cmidrule(lr){1-1} \cmidrule(lr){3-22}
MT\tiny{(NeurIPS'17)~\cite{tarvainen2017mean}} & & 90.23\tiny{$\pm$10.05} & 84.00\tiny{$\pm$12.85} & 6.65\tiny{$\pm$9.56} & 2.09\tiny{$\pm$2.98} & 79.77\tiny{$\pm$20.48} & 70.62\tiny{$\pm$22.58} & 21.59\tiny{$\pm$29.31} & 7.92\tiny{$\pm$11.70} & 89.17\tiny{$\pm$14.84} & 82.99\tiny{$\pm$17.03} & 6.71\tiny{$\pm$10.62} & 2.03\tiny{$\pm$3.25} & 76.92\tiny{$\pm$24.65} & 68.03\tiny{$\pm$25.64} & 16.49\tiny{$\pm$24.93} & 5.43\tiny{$\pm$9.29} & 84.02\tiny{$\pm$20.28} & 76.41\tiny{$\pm$22.09} & 12.86\tiny{$\pm$22.86} & 4.37\tiny{$\pm$8.88}  \\
UAMT\tiny{(MICCAI'19)~\cite{yu2019uncertainty}}  & & 89.26\tiny{$\pm$13.78} & 82.72\tiny{$\pm$17.07} & 8.32\tiny{$\pm$11.67} & 2.56\tiny{$\pm$3.72} & 78.93\tiny{$\pm$23.31} & 70.38\tiny{$\pm$24.81} & 22.02\tiny{$\pm$31.79} & 8.11\tiny{$\pm$13.04} & 88.27\tiny{$\pm$16.89} & 81.99\tiny{$\pm$19.33} & 7.97\tiny{$\pm$11.55} & 2.36\tiny{$\pm$3.47} & 76.21\tiny{$\pm$28.95} & 68.54\tiny{$\pm$29.12} & 16.13\tiny{$\pm$25.21} & 5.49\tiny{$\pm$10.19} & 83.16\tiny{$\pm$22.76} & 75.91\tiny{$\pm$24.28} & 13.61\tiny{$\pm$23.20} & 4.63\tiny{$\pm$9.27} \\
URPC\tiny{(MIA'22)~\cite{luo2022semi}}  & & 88.28\tiny{$\pm$15.15} & 82.14\tiny{$\pm$16.76} & 4.00\tiny{$\pm$8.34} & 0.97\tiny{$\pm$2.23} & 80.47\tiny{$\pm$22.35} & 71.91\tiny{$\pm$23.71} & 14.61\tiny{$\pm$25.01} & 4.94\tiny{$\pm$9.69} & 87.97\tiny{$\pm$15.07} & 82.09\tiny{$\pm$16.54} & 3.80\tiny{$\pm$8.26} & 0.84\tiny{$\pm$2.31} & 78.28\tiny{$\pm$24.91} & 70.66\tiny{$\pm$25.48} & 11.83\tiny{$\pm$24.52} & 4.04\tiny{$\pm$10.09} & 83.75\tiny{$\pm$21.51} & 76.70\tiny{$\pm$22.59} & 8.56\tiny{$\pm$19.99} & 2.70\tiny{$\pm$7.76} \\
BCP\tiny{(CVPR'23)~\cite{bai2023bidirectional}}  & & 93.07\tiny{$\pm$2.51} & 87.16\tiny{$\pm$4.18} & 2.27\tiny{$\pm$1.80} & 0.86\tiny{$\pm$0.40} & 90.17\tiny{$\pm$5.09} & 82.58\tiny{$\pm$7.25} & 4.63\tiny{$\pm$8.27} & 1.46\tiny{$\pm$2.79} & 92.39\tiny{$\pm$3.75} & 86.12\tiny{$\pm$5.87} & 2.35\tiny{$\pm$1.19} & 0.93\tiny{$\pm$0.43} & 90.66\tiny{$\pm$6.59} & 83.70\tiny{$\pm$8.62} & 5.84\tiny{$\pm$14.17} & 1.64\tiny{$\pm$3.59} & 91.57\tiny{$\pm$5.26} & 84.89\tiny{$\pm$7.41} & 3.77\tiny{$\pm$8.91} & 1.22\tiny{$\pm$2.48}  \\
ABD\tiny{(CVPR'24)~\cite{chi2024adaptive}}   & & \underline{95.28\tiny{$\pm$0.83}} & \underline{91.08\tiny{$\pm$1.20}} & 2.13\tiny{$\pm$1.80} & 0.68\tiny{$\pm$0.63} & \underline{90.91\tiny{$\pm$1.07}} & \underline{83.67\tiny{$\pm$1.19}} & 3.29\tiny{$\pm$4.37} & 1.08\tiny{$\pm$1.82} & \underline{95.33\tiny{$\pm$1.47}} & \underline{91.23\tiny{$\pm$1.84}} & \underline{1.88\tiny{$\pm$1.36}} & 0.62\tiny{$\pm$0.43} & \underline{91.07\tiny{$\pm$3.05}} & \underline{84.19\tiny{$\pm$2.32}} & 3.32\tiny{$\pm$6.06} & 1.07\tiny{$\pm$2.54} & \underline{93.15\tiny{$\pm$1.44}} & \underline{87.54\tiny{$\pm$1.26}} & 2.65\tiny{$\pm$2.81} & 0.86\tiny{$\pm$1.01} \\ \cmidrule(lr){1-1} \cmidrule(lr){3-22}
CPC-SAM\tiny{(MICCAI'24)~\cite{miao2024cross}} & & 95.25\tiny{$\pm$1.27} & 90.95\tiny{$\pm$2.28} & \textbf{1.90\tiny{$\pm$0.60}} & \textbf{0.58\tiny{$\pm$0.17}} & 89.11\tiny{$\pm$3.74} & 80.56\tiny{$\pm$5.76} & \underline{2.66\tiny{$\pm$0.67}} & \underline{0.97\tiny{$\pm$0.77}} & 95.05\tiny{$\pm$1.46} & 90.61\tiny{$\pm$2.63} & 1.92\tiny{$\pm$0.59} & \underline{0.59\tiny{$\pm$0.18}} & 90.19\tiny{$\pm$2.68} & 82.23\tiny{$\pm$4.33} & \underline{2.54\tiny{$\pm$0.66}} & \underline{0.82\tiny{$\pm$0.19}} & 92.40\tiny{$\pm$3.73} & 86.09\tiny{$\pm$6.20} & \underline{2.25\tiny{$\pm$0.72}} & \underline{0.74\tiny{$\pm$0.45}} \\
\cellcolor{lightblue!50}\textbf{UnCoL(ours)} & & \cellcolor{lightblue!50}\textbf{95.63\tiny{$\pm$1.13}} & \cellcolor{lightblue!50}\textbf{91.64\tiny{$\pm$2.02}} & \cellcolor{lightblue!50}\underline{2.07\tiny{$\pm$4.63}} & \cellcolor{lightblue!50}\underline{0.66\tiny{$\pm$1.03}} & \cellcolor{lightblue!50}\textbf{93.01\tiny{$\pm$2.34}} & \cellcolor{lightblue!50}\textbf{87.03\tiny{$\pm$3.94}} & \cellcolor{lightblue!50}\textbf{2.13\tiny{$\pm$2.17}} & \cellcolor{lightblue!50}\textbf{0.57\tiny{$\pm$0.45}} & \cellcolor{lightblue!50}\textbf{95.73\tiny{$\pm$1.07}} & \cellcolor{lightblue!50}\textbf{91.83\tiny{$\pm$1.93}} & \cellcolor{lightblue!50}\textbf{1.37\tiny{$\pm$0.45}} & \cellcolor{lightblue!50}\textbf{0.50\tiny{$\pm$0.14}} & \cellcolor{lightblue!50}\textbf{93.32\tiny{$\pm$2.12}} & \cellcolor{lightblue!50}\textbf{87.56\tiny{$\pm$3.63}} & \cellcolor{lightblue!50}\textbf{2.69\tiny{$\pm$4.62}} & \cellcolor{lightblue!50}\textbf{0.73\tiny{$\pm$0.89}} & \cellcolor{lightblue!50}\textbf{94.42\tiny{$\pm$2.17}} & \cellcolor{lightblue!50}\textbf{89.51\tiny{$\pm$3.77}} & \cellcolor{lightblue!50}\textbf{2.06\tiny{$\pm$4.07}} & \cellcolor{lightblue!50}\textbf{0.61\tiny{$\pm$0.87}} \\ \midrule\midrule
UNet\tiny{(MICCAI'15)~\cite{ronneberger2015u}}  &\multirow{2}{*}{\makecell{{289/289}\\{($100\%$)}}}  &94.34\tiny{$\pm$8.80} & 90.38\tiny{$\pm$9.79} & 2.80\tiny{$\pm$5.99} & 0.95\tiny{$\pm$2.16} & 88.69\tiny{$\pm$16.05} & 82.23\tiny{$\pm$17.16} & 4.59\tiny{$\pm$11.61} & 1.31\tiny{$\pm$3.94} & 94.71\tiny{$\pm$6.91} & 90.69\tiny{$\pm$8.50} & 2.61\tiny{$\pm$5.65} & 0.91\tiny{$\pm$1.95} & 91.05\tiny{$\pm$10.24} & 84.82\tiny{$\pm$12.38} & 11.79\tiny{$\pm$24.31} & 3.87\tiny{$\pm$8.62} & 92.20\tiny{$\pm$11.90} & 87.03\tiny{$\pm$13.41} & 5.45\tiny{$\pm$15.14} & 1.76\tiny{$\pm$5.32}  \\
{nnU-Net\tiny{(Nat. Methods'21)~\cite{isensee2021nnu}}} &  & {95.05\tiny{$\pm$ 7.56}} & {91.30\tiny{$\pm$ 7.86}} & {1.88\tiny{$\pm$ 3.77}} & {0.71\tiny{$\pm$ 1.83}} & {92.65\tiny{$\pm$ 6.55}} & {86.87\tiny{$\pm$ 7.73}} & {2.96\tiny{$\pm$ 7.47}} & {1.10\tiny{$\pm$ 4.08}} & {95.09\tiny{$\pm$ 7.07}} & {91.30\tiny{$\pm$ 7.54}} & {1.80\tiny{$\pm$ 3.28}} & {0.69\tiny{$\pm$ 1.54}} & {93.07\tiny{$\pm$ 6.38}} & {87.57\tiny{$\pm$ 7.78}} & {3.07\tiny{$\pm$ 8.06}} & {1.17\tiny{$\pm$ 4.30}} & {93.96\tiny{$\pm$ 7.13}} & {89.26\tiny{$\pm$ 8.12}} & {2.43\tiny{$\pm$ 6.15}} & {0.92\tiny{$\pm$ 3.27}} \\
\bottomrule[1.3pt]
\end{tabular}}
\end{table*}

\begin{figure*}[htbp]
    \centering
    \includegraphics[width=0.98\textwidth]{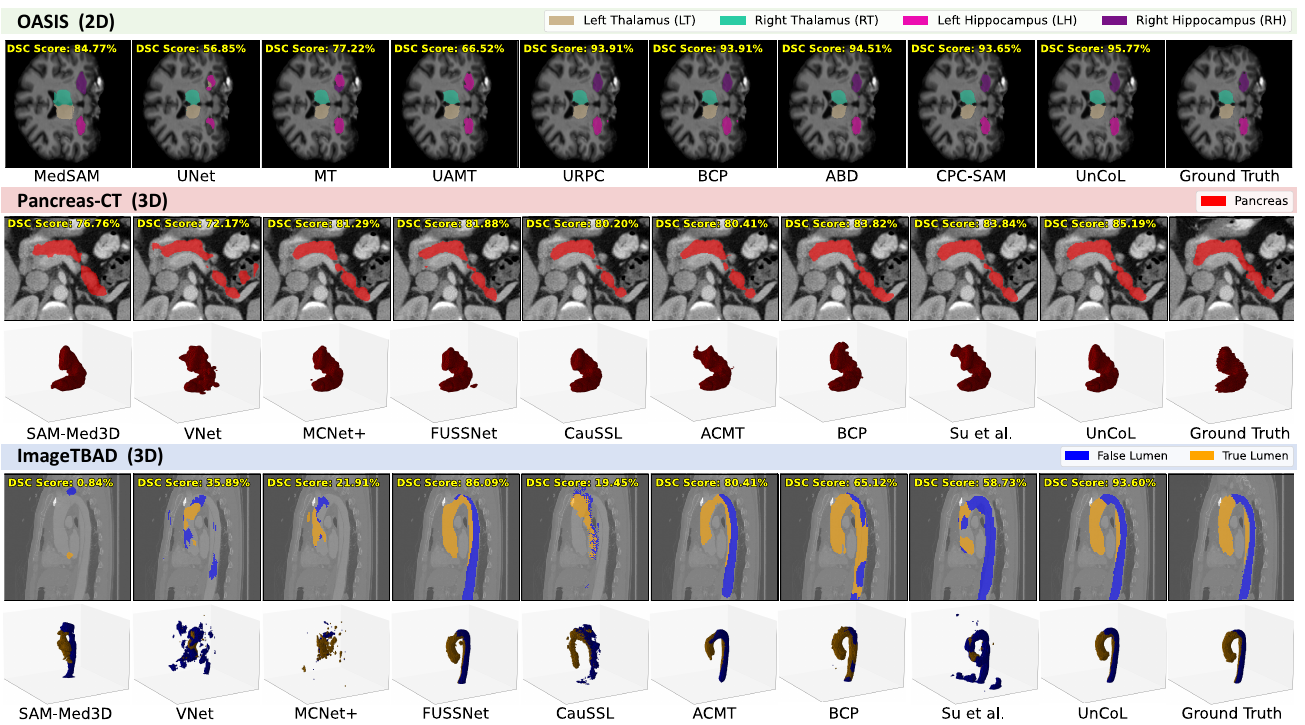}
    \caption{
    Visualization of segmentation results. Each row shows the predictions (color-coded by class) with corresponding Dice scores, and the last column shows the ground truth. The first column illustrates zero-shot outputs from foundation models, while the rest show results from models trained with partial labels (5\% for OASIS; 20\% for Pancreas-CT and ImageTBAD).
    }
    \label{fig:tbad}
\end{figure*}

\begin{table}[htbp]
\centering
\caption{Performance comparison on the Pancreas-CT dataset. 
{``Labeled Ratio" indicates the ratio of labeled data.}
95HD and ASD are reported in voxel units.
\textbf{Bold} and \underline{underline} indicate the best and second-best results.
} \label{tab:pancreas}
\resizebox{0.48\textwidth}{!}{
\begin{tabular}{l|c|llll}
\toprule[1.3pt]
\multicolumn{1}{c|}{\multirow{2}{*}{Method}} & \multicolumn{1}{c}{\multirow{2}{*}{\makecell{{Labeled}\\{Ratio}}}} & \multicolumn{4}{|c}{Metrics}  \\ \cmidrule{3-6}
\multicolumn{1}{c|}{}  &  & DSC$(\%)\uparrow$ & Jaccard$(\%)\uparrow$ & 95HD$\downarrow$   & ASD$\downarrow$ \\ \midrule\midrule
SAM\tiny{(ICCV'23)~\cite{kirillov2023segment}}  & \multirow{3}{*}{\makecell{{0/62 ($0\%$)}}} & 18.13\tiny{$\pm$}7.37 & 10.16\tiny{$\pm$}4.50 & 17.86\tiny{$\pm$}6.57 & 4.27\tiny{$\pm$}1.80\\
MedSAM\tiny{(NC'24)~\cite{ma2024segment}} & & 70.92\tiny{$\pm$}4.08 & 55.10\tiny{$\pm$}4.82 & 7.71\tiny{$\pm$}1.59 & 2.68\tiny{$\pm$}0.51  \\
SAM-Med3D\tiny{~\cite{wang2023sam}} & & 79.37\tiny{$\pm$}4.55 & 66.02\tiny{$\pm$}6.22 & 5.52\tiny{$\pm$}1.79 & 1.55\tiny{$\pm$}0.54\\ 
\midrule\midrule
VNet\tiny{(3DV'16)~\cite{milletari2016v}} & \multirow{13}{*}{\makecell{{6/62 ($\sim10\%$)}}} & 56.96\tiny{$\pm$16.50} & 41.38\tiny{$\pm$13.66} & 18.69\tiny{$\pm$12.60} & 5.38\tiny{$\pm$3.90} \\
{nnU-Net\tiny{(Nat. Methods'21)~\cite{isensee2021nnu}}} & & {49.06\tiny{$\pm$ 20.36}} & {34.93\tiny{$\pm$ 16.73}} & {23.66\tiny{$\pm$ 10.13}} & {8.62\tiny{$\pm$ 4.87}} \\
{\scriptsize{Swin-UNETR}\tiny{(MICCAI'21)~\cite{hatamizadeh2021swin}}} & & {67.11\tiny{$\pm$10.59}} & {51.42\tiny{$\pm$11.51}} & {12.35\tiny{$\pm$11.99}} & {2.91\tiny{$\pm$3.10}} \\ \cmidrule(lr){1-1} \cmidrule(lr){3-6}
MT\tiny{(NeurIPS’17)~\cite{tarvainen2017mean}} & & 64.28\tiny{$\pm$11.62} & 48.40\tiny{$\pm$12.10} & 28.19\tiny{$\pm$14.54} & 8.92\tiny{$\pm$5.15}  \\
UA-MT\tiny{(MICCAI'19)~\cite{yu2019uncertainty}} & & 62.04\tiny{$\pm$8.07} & 45.46\tiny{$\pm$8.44} & 33.54\tiny{$\pm$15.97} & 11.26\tiny{$\pm$5.22} \\
MC-Net+\tiny{(MIA'22)~\cite{wu2022mutual}}  & & 64.26\tiny{$\pm$13.61} & 48.82\tiny{$\pm$14.82} & 33.63\tiny{$\pm$23.19} & 13.02\tiny{$\pm$9.57} \\
FUSSNet\tiny{(MICCAI'22)~\cite{xiang2022fussnet}}  &  & 59.32\tiny{$\pm$12.50} & 43.24\tiny{$\pm$12.25} & 30.83\tiny{$\pm$13.64} & 10.29\tiny{$\pm$5.94} \\
CauSSL\tiny{(ICCV'23)~\cite{miao2023caussl}}   & & 64.35\tiny{$\pm$13.30} & 48.75\tiny{$\pm$13.54} & 23.01\tiny{$\pm$11.60}  & 7.77\tiny{$\pm$4.13} \\
AC-MT\tiny{(MIA'23)~\cite{xu2023ambiguity}} & & 66.50\tiny{$\pm$8.72} & 50.43\tiny{$\pm$9.42} & 21.96\tiny{$\pm$11.48} & 7.31\tiny{$\pm$3.90} \\
BCP\tiny{(CVPR'23)~\cite{bai2023bidirectional}} & & \textbf{80.73\tiny{$\pm$8.29}} & \textbf{68.48\tiny{$\pm$10.17}} & \underline{7.67\tiny{$\pm$7.35}} & \underline{2.38\tiny{$\pm$2.41}} \\
Su et al.\tiny{(MIA'24)~\cite{su2024mutual}} & & 71.29\tiny{$\pm$9.70} & 56.25\tiny{$\pm$11.50} & 18.19\tiny{$\pm$12.57} & 5.97\tiny{$\pm$4.45} \\ \cmidrule(lr){1-1} \cmidrule(lr){3-6}
SemiSAM+\tiny{(arXiv'25)~\cite{zhang2025semisam+}} & & 64.72\tiny{$\pm$10.33} & 48.65\tiny{$\pm$10.52} & 20.53\tiny{$\pm$10.39} & 6.10\tiny{$\pm$3.19}  \\
\cellcolor{lightblue!50}\textbf{UnCoL(ours)} & & \cellcolor{lightblue!50}\underline{80.40\tiny{$\pm$5.82}} & \cellcolor{lightblue!50}\underline{67.60\tiny{$\pm$7.83}} & \cellcolor{lightblue!50}\textbf{6.86\tiny{$\pm$3.96}} & \cellcolor{lightblue!50}\textbf{1.85\tiny{$\pm$0.94}} \\ \midrule\midrule
VNet\tiny{(3DV'16)~\cite{milletari2016v}} & \multirow{13}{*}{\makecell{{12/62 ($\sim20\%$)}}} & 70.52\tiny{$\pm$11.40} & 55.62\tiny{$\pm$13.12} & 10.43\tiny{$\pm$6.05} & 2.28\tiny{$\pm$1.83} \\
{nnU-Net\tiny{(Nat. Methods'21)~\cite{isensee2021nnu}}} & & {60.06\tiny{$\pm$ 17.06}} & {45.08\tiny{$\pm$ 15.95}} & {16.65\tiny{$\pm$ 9.14}} & {5.51\tiny{$\pm$ 4.04}} \\
{\scriptsize{Swin-UNETR}\tiny{(MICCAI'21)~\cite{hatamizadeh2021swin}}} & & {75.42\tiny{$\pm$7.73}} & {61.14\tiny{$\pm$9.62}} & {5.12\tiny{$\pm$3.52}} & {1.00\tiny{$\pm$0.98}} \\ \cmidrule(lr){1-1} \cmidrule(lr){3-6}
MT\tiny{(NeurIPS’17)~\cite{tarvainen2017mean}} & & 74.97\tiny{$\pm$8.21} & 60.63\tiny{$\pm$10.17} & 12.18\tiny{$\pm$8.72} & 3.59\tiny{$\pm$2.56} \\
UA-MT\tiny{(MICCAI'19)~\cite{yu2019uncertainty}} & & 76.45\tiny{$\pm$7.97} & 62.53\tiny{$\pm$10.12} & 9.47\tiny{$\pm$5.25} & 2.99\tiny{$\pm$1.54} \\
MC-Net+\tiny{(MIA'22)~\cite{wu2022mutual}}  & & 77.98\tiny{$\pm$8.76} & 64.69\tiny{$\pm$10.95} & 8.97\tiny{$\pm$6.50} & 2.63\tiny{$\pm$1.75} \\
FUSSNet\tiny{(MICCAI'22)~\cite{xiang2022fussnet}}   &  & 77.26\tiny{$\pm$8.15} & 63.64\tiny{$\pm$10.56} & 12.66\tiny{$\pm$9.57} & 4.02\tiny{$\pm$3.09}  \\
CauSSL\tiny{(ICCV'23)~\cite{miao2023caussl}}   & & 77.44\tiny{$\pm$9.27} & 64.02\tiny{$\pm$11.17} & 11.96\tiny{$\pm$10.49} & 3.52\tiny{$\pm$3.10} \\
AC-MT\tiny{(MIA'23)~\cite{xu2023ambiguity}} & & 76.39\tiny{$\pm$8.63} & 62.55\tiny{$\pm$10.78} & 10.80\tiny{$\pm$7.04} & 3.30\tiny{$\pm$2.23}  \\
BCP\tiny{(CVPR'23)~\cite{bai2023bidirectional}} & & \underline{82.93\tiny{$\pm$5.56}} & \underline{71.21\tiny{$\pm$7.65}} & \underline{6.30\tiny{$\pm$5.46}} & 1.83\tiny{$\pm$1.34}  \\
Su et al.\tiny{(MIA'24)~\cite{su2024mutual}} & & 81.01\tiny{$\pm$6.04} & 68.50\tiny{$\pm$8.26 } & 6.32\tiny{$\pm$3.28} & \underline{1.72\tiny{$\pm$1.14}} \\ \cmidrule(lr){1-1} \cmidrule(lr){3-6}
SemiSAM+\tiny{(arXiv'25)~\cite{zhang2025semisam+}} & & 75.97\tiny{$\pm$6.96} & 61.76\tiny{$\pm$8.97} & 8.89\tiny{$\pm$5.76} & 2.57\tiny{$\pm$1.37} \\
\cellcolor{lightblue!50}\textbf{UnCoL(ours)} & & \cellcolor{lightblue!50}\textbf{83.16\tiny{$\pm$4.80}} & \cellcolor{lightblue!50}\textbf{71.46\tiny{$\pm$6.79}} & \cellcolor{lightblue!50}\textbf{5.44\tiny{$\pm$3.26}} & \cellcolor{lightblue!50}\textbf{1.44\tiny{$\pm$0.90}} \\ \midrule\midrule
VNet\tiny{(3DV'16)~\cite{milletari2016v}} & \multirow{4}{*}{\makecell{{62/62 ($100\%$)}}}& 81.89\tiny{$\pm$6.33} & 69.78\tiny{$\pm$8.46} & 9.33\tiny{$\pm$8.26} & 2.39\tiny{$\pm$1.99} \\
{nnU-Net\tiny{(Nat. Methods'21)~\cite{isensee2021nnu}}} & & {79.93\tiny{$\pm$ 7.05}} & {67.91\tiny{$\pm$ 8.72}} & {6.32\tiny{$\pm$ 3.35}} & {1.94\tiny{$\pm$ 0.89}} \\
{\scriptsize{Swin-UNETR}\tiny{(MICCAI'21)~\cite{hatamizadeh2021swin}}} & & {83.49\tiny{$\pm$3.98}} & {71.86\tiny{$\pm$5.75}} & {2.35\tiny{$\pm$1.91}} & {0.45\tiny{$\pm$0.32}}  \\
{SegVol\tiny{(NeurIPS'24)~\cite{du2024segvol}}} & & {58.83\tiny{$\pm$16.46}} & {42.23\tiny{$\pm$16.54}} & {21.62\tiny{$\pm$19.64}} & {5.43\tiny{$\pm$4.85}} \\
\bottomrule[1.3pt]
\end{tabular}}
\end{table}

\begin{table*}[htbp]
\centering
\caption{Performance comparison on the ImageTBAD dataset. 
{``Labeled Ratio" indicates the ratio of labeled data.}
95HD and ASD are reported in voxel units.
\textbf{Bold} and \underline{underline} indicate the best and second-best results.
} \label{tab:tbad}
\resizebox{0.98\textwidth}{!}{
\begin{tabular}{l|c|llll|llll|llll}
\toprule[1.3pt]
\multicolumn{1}{c}{\multirow{2}{*}{Methods}} & \multicolumn{1}{|c}{\multirow{2}{*}{\makecell{{Labeled}\\{Ratio}}}} & \multicolumn{4}{|c}{True Lumen (TL)}   & \multicolumn{4}{|c|}{False Lumen (FL)} & \multicolumn{4}{c}{Average}  \\ \cmidrule{3-14}
\multicolumn{1}{c}{} & \multicolumn{1}{|c|}{}  & \multicolumn{1}{c}{DSC\tiny{$(\%)$}\normalsize{$\uparrow$}} & Jaccard\tiny{$(\%)$}\normalsize{$\uparrow$} & 95HD$\downarrow$   & ASD$\downarrow$  & DSC\tiny{$(\%)$}\normalsize{$\uparrow$} & Jaccard\tiny{$(\%)$}\normalsize{$\uparrow$} & 95HD$\downarrow$   & ASD$\downarrow$  & DSC\tiny{$(\%)$}\normalsize{$\uparrow$} & Jaccard\tiny{$(\%)$}\normalsize{$\uparrow$} & 95HD$\downarrow$   & ASD$\downarrow$  \\ \midrule\midrule
SAM\tiny{(ICCV'23)~\cite{kirillov2023segment}}   & \multirow{4}{*}{\makecell{{0/70}\\{($0\%$)}}} & 19.18\tiny{$\pm$}4.48 & 10.99\tiny{$\pm$}2.89 & 16.14\tiny{$\pm$}7.61 & 5.64\tiny{$\pm$}1.30 & 18.17\tiny{$\pm$}3.53 & 10.31\tiny{$\pm$}2.25 & 16.42\tiny{$\pm$}3.98 & 5.55\tiny{$\pm$}1.03 & 18.68\tiny{$\pm$}4.07 & 10.65\tiny{$\pm$}2.62 & 16.28\tiny{$\pm$}6.15 & 5.60\tiny{$\pm$}1.18 \\
MedSAM\tiny{(NC'24)~\cite{ma2024segment}}  &  & 47.88\tiny{$\pm$}11.68 & 32.23\tiny{$\pm$}9.70 & 7.89\tiny{$\pm$}2.75 & 2.89\tiny{$\pm$}0.95 & 40.35\tiny{$\pm$}6.21 & 25.48\tiny{$\pm$}4.91 & 11.73\tiny{$\pm$}2.43 & 4.54\tiny{$\pm$}0.81 & 44.12\tiny{$\pm$}10.13 & 28.86\tiny{$\pm$}8.43 & 9.81\tiny{$\pm$}3.26 & 3.71\tiny{$\pm$}1.22  \\
SAM-Med3D\tiny{~\cite{wang2023sam}} &  & 33.56\tiny{$\pm$}20.60 & 22.23\tiny{$\pm$}15.73 & 13.55\tiny{$\pm$}6.71 & 4.42\tiny{$\pm$}2.52 & 25.66\tiny{$\pm$}16.18 & 15.86\tiny{$\pm$}11.85 & 12.91\tiny{$\pm$}3.93 & 4.47\tiny{$\pm$}1.06 & 29.61\tiny{$\pm$}19.11 & 19.04\tiny{$\pm$}14.43 & 13.23\tiny{$\pm$}5.58 & 4.44\tiny{$\pm$}1.94 \\ 
{SegVol\tiny{(NeurIPS'24)~\cite{du2024segvol}}} & & {34.41\tiny{$\pm$15.24} } & {21.85\tiny{$\pm$11.73} } & {18.68\tiny{$\pm$7.39} } & {5.90\tiny{$\pm$2.93} } & {30.87\tiny{$\pm$16.84} } & {19.22\tiny{$\pm$12.29} } & {19.64\tiny{$\pm$10.90} } & {6.01\tiny{$\pm$3.37} } & {32.64\tiny{$\pm$16.15} } & {20.53\tiny{$\pm$12.09} } & {19.16\tiny{$\pm$9.32} } & {5.96\tiny{$\pm$3.16}} \\ \midrule\midrule
VNet\tiny{(3DV'16)~\cite{milletari2016v}}  & \multirow{10}{*}{\makecell{{7/70}\\{($10\%$)}}} &  63.25\tiny{$\pm$14.37} & 48.41\tiny{$\pm$14.05} & 13.34\tiny{$\pm$10.49} & 3.20\tiny{$\pm$4.00} & 48.45\tiny{$\pm$23.80} & 36.41\tiny{$\pm$20.80} & 15.04\tiny{$\pm$12.32} & 4.09\tiny{$\pm$5.00} & 55.85\tiny{$\pm$21.20} & 42.41\tiny{$\pm$18.96} & 14.19\tiny{$\pm$11.74} & 3.65\tiny{$\pm$4.56} \\
{nnU-Net\tiny{(Nat. Methods'21)~\cite{isensee2021nnu}}} &  & {73.14\tiny{$\pm$ 11.63}} & {59.04\tiny{$\pm$ 13.44}} & {6.25\tiny{$\pm$ 3.96}} & {1.67\tiny{$\pm$ 1.03}} & {64.07\tiny{$\pm$ 20.09}} & {50.34\tiny{$\pm$ 20.28}} & {8.10\tiny{$\pm$ 9.53}} & {2.49\tiny{$\pm$ 2.94}} & {68.61\tiny{$\pm$ 17.28}} & {54.69\tiny{$\pm$ 18.04}} & {7.17\tiny{$\pm$ 8.04}} & {2.08\tiny{$\pm$ 2.26}} \\ 
{\scriptsize{Swin-UNETR}\tiny{(MICCAI'21)~\cite{hatamizadeh2021swin}}} &  & {64.74\tiny{$\pm$ 9.35}} & {48.74\tiny{$\pm$ 9.81}} & {6.02\tiny{$\pm$ 5.25}} & {1.75\tiny{$\pm$ 2.22}} & {59.74\tiny{$\pm$ 13.70}} & {44.04\tiny{$\pm$ 13.64}} & {5.85\tiny{$\pm$ 6.51}} & {1.57\tiny{$\pm$ 1.72}} & {62.24\tiny{$\pm$ 12.28}} & {46.39\tiny{$\pm$ 12.44}} & {5.94\tiny{$\pm$ 6.33}} & {1.66\tiny{$\pm$ 2.16}} \\  \cmidrule(lr){1-1} \cmidrule(lr){3-14}
MT\tiny{(NeurIPS’17)~\cite{tarvainen2017mean}}  &  & 70.41\tiny{$\pm$9.54} & 55.40\tiny{$\pm$10.91} & 5.89\tiny{$\pm$4.75} & 1.46\tiny{$\pm$0.79} & 58.92\tiny{$\pm$18.60} & 44.68\tiny{$\pm$17.82} & 9.07\tiny{$\pm$4.57} & 2.54\tiny{$\pm$1.58} & 64.67\tiny{$\pm$15.93} & 50.04\tiny{$\pm$15.83} & 7.48\tiny{$\pm$6.06} & 2.00\tiny{$\pm$1.43} \\
UA-MT\tiny{(MICCAI'19)~\cite{yu2019uncertainty}} &  & 74.21\tiny{$\pm$8.90} & 59.77\tiny{$\pm$10.98} & 4.78\tiny{$\pm$2.11} & \underline{1.27\tiny{$\pm$0.60}} & 63.14\tiny{$\pm$20.28} & 49.45\tiny{$\pm$20.52} & 5.47\tiny{$\pm$4.19} & 1.50\tiny{$\pm$1.12} & 68.67\tiny{$\pm$16.77} & 54.61\tiny{$\pm$17.41} & \underline{5.12\tiny{$\pm$3.35}} & \underline{1.39\tiny{$\pm$0.91}}  \\
MC-Net+\tiny{(MIA'22)~\cite{wu2022mutual}}   & & 61.80\tiny{$\pm$17.04} & 47.02\tiny{$\pm$16.40} & 9.96\tiny{$\pm$10.09} & 2.78\tiny{$\pm$2.81} & 52.10\tiny{$\pm$22.80} & 38.67\tiny{$\pm$20.51} & 10.85\tiny{$\pm$11.09} & 4.04\tiny{$\pm$4.88} & 56.95\tiny{$\pm$20.87} & 42.85\tiny{$\pm$19.21} & 10.40\tiny{$\pm$10.75} & 3.41\tiny{$\pm$4.14}  \\
FUSSNet\tiny{(MICCAI'22)~\cite{xiang2022fussnet}}   & & 73.95\tiny{$\pm$11.23} & 60.48\tiny{$\pm$12.96} & 6.08\tiny{$\pm$3.56} & 1.49\tiny{$\pm$1.13} & \underline{63.84\tiny{$\pm$18.71}} & \underline{49.84\tiny{$\pm$19.32}} & 6.29\tiny{$\pm$6.89} & 1.86\tiny{$\pm$1.76} & \underline{68.90\tiny{$\pm$16.59}} & \underline{55.16\tiny{$\pm$17.55}} & 6.19\tiny{$\pm$5.65} & 1.67\tiny{$\pm$1.53} \\
CauSSL\tiny{(ICCV'23)~\cite{miao2023caussl}}   & & 51.61\tiny{$\pm$14.95} & 36.26\tiny{$\pm$13.41} & 7.90\tiny{$\pm$5.24} & 2.20\tiny{$\pm$0.80} & 33.14\tiny{$\pm$13.95} & 22.94\tiny{$\pm$11.44} & 8.07\tiny{$\pm$10.43} & 2.24\tiny{$\pm$2.74} & 42.37\tiny{$\pm$20.10} & 29.60\tiny{$\pm$16.39} & 7.98\tiny{$\pm$9.05} & 2.22\tiny{$\pm$2.11} \\
AC-MT\tiny{(MIA'23)~\cite{xu2023ambiguity}}   &  & 72.80\tiny{$\pm$8.77} & 58.09\tiny{$\pm$10.67} & 5.21\tiny{$\pm$3.30} & 1.38\tiny{$\pm$0.54} & 62.61\tiny{$\pm$17.17} & 47.96\tiny{$\pm$17.57} & \underline{5.42\tiny{$\pm$4.68}} & 1.64\tiny{$\pm$1.29} & 67.70\tiny{$\pm$14.65} & 53.03\tiny{$\pm$15.53} & 5.32\tiny{$\pm$4.27} & 1.51\tiny{$\pm$1.02} \\
BCP\tiny{(CVPR'23)~\cite{bai2023bidirectional}} &  & \underline{76.06\tiny{$\pm$7.92}} & \underline{62.01\tiny{$\pm$10.07}} & \underline{4.56\tiny{$\pm$2.30}} & 1.46\tiny{$\pm$0.73} & 60.86\tiny{$\pm$20.50} & 46.72\tiny{$\pm$20.33} & 5.94\tiny{$\pm$7.77} & 1.59\tiny{$\pm$2.32} & 68.46\tiny{$\pm$17.29} & 54.36\tiny{$\pm$17.77} & 5.25\tiny{$\pm$5.77} & 1.53\tiny{$\pm$1.72} \\
Su et al.\tiny{(MIA'24)~\cite{su2024mutual}} & & 69.43\tiny{$\pm$11.74} & 54.50\tiny{$\pm$12.30} & 8.46\tiny{$\pm$7.17} & 2.44\tiny{$\pm$1.87} & 63.25\tiny{$\pm$15.45} & 48.20\tiny{$\pm$16.36} & 7.38\tiny{$\pm$8.22} & \textbf{1.35\tiny{$\pm$0.75}} & 66.34\tiny{$\pm$14.50} & 51.35\tiny{$\pm$15.04} & 7.92\tiny{$\pm$7.95} & 1.90\tiny{$\pm$1.56}  \\ \cmidrule(lr){1-1} \cmidrule(lr){3-14}
SemiSAM+\tiny{(arXiv'25)~\cite{zhang2025semisam+}} & & 60.46\tiny{$\pm$10.66} & 44.17\tiny{$\pm$11.07} & 21.64\tiny{$\pm$14.04} & 5.88\tiny{$\pm$3.39} & 61.65\tiny{$\pm$13.90} & 45.92\tiny{$\pm$13.64} & 21.58\tiny{$\pm$14.12} & 5.68\tiny{$\pm$3.05} & 61.05\tiny{$\pm$12.40} & 45.05\tiny{$\pm$12.45} & 21.61\tiny{$\pm$14.08} & 5.78\tiny{$\pm$3.22} \\
\cellcolor{lightblue!50}\textbf{UnCoL(ours)}  & & \cellcolor{lightblue!50}\textbf{77.57\tiny{$\pm$7.31}} & \cellcolor{lightblue!50}\textbf{64.33\tiny{$\pm$9.34}} & \cellcolor{lightblue!50}\textbf{4.24\tiny{$\pm$1.63}} & \cellcolor{lightblue!50}\textbf{1.18\tiny{$\pm$0.42}} & \cellcolor{lightblue!50}\textbf{64.37\tiny{$\pm$14.03}} & \cellcolor{lightblue!50}\textbf{50.01\tiny{$\pm$14.02}} & \cellcolor{lightblue!50}\textbf{5.41\tiny{$\pm$3.90}} & \cellcolor{lightblue!50}\underline{1.42\tiny{$\pm$1.16}} & \cellcolor{lightblue!50}\textbf{70.97\tiny{$\pm$12.58}} & \cellcolor{lightblue!50}\textbf{57.17\tiny{$\pm$13.45}} & \cellcolor{lightblue!50}\textbf{4.83\tiny{$\pm$3.16}} & \cellcolor{lightblue!50}\textbf{1.30\tiny{$\pm$0.90}} \\ \midrule\midrule
VNet\tiny{(3DV'16)~\cite{milletari2016v}}  & \multirow{10}{*}{\makecell{{14/70}\\{($20\%$)}}} &  70.12\tiny{$\pm$11.90} & 55.35\tiny{$\pm$13.74} & 14.59\tiny{$\pm$15.45} & 4.04\tiny{$\pm$5.06} & 51.95\tiny{$\pm$26.63} & 39.70\tiny{$\pm$23.01} & 14.33\tiny{$\pm$15.91} & 6.15\tiny{$\pm$7.92} & 61.04\tiny{$\pm$22.62} & 47.52\tiny{$\pm$20.63} & 14.46\tiny{$\pm$15.70} & 5.11\tiny{$\pm$6.75} \\
{nnU-Net\tiny{(Nat. Methods'21)~\cite{isensee2021nnu}}} &  & {76.21\tiny{$\pm$ 8.92}} & {62.48\tiny{$\pm$ 11.16}} & {5.95\tiny{$\pm$ 4.24}} & {1.67\tiny{$\pm$ 1.20}} & {67.99\tiny{$\pm$ 16.11}} & {53.95\tiny{$\pm$ 17.20}} & {6.99\tiny{$\pm$ 8.43}} & {1.97\tiny{$\pm$ 1.97}} & {72.10\tiny{$\pm$ 13.95}} & {58.21\tiny{$\pm$ 15.51}} & {6.47\tiny{$\pm$ 7.31}} & {1.81\tiny{$\pm$ 1.71}} \\ 
{\scriptsize{Swin-UNETR}\tiny{(MICCAI'21)~\cite{hatamizadeh2021swin}}} &  & {70.12\tiny{$\pm$ 8.80}} & {54.75\tiny{$\pm$ 9.86}} & {3.63\tiny{$\pm$ 1.78}} & {0.91\tiny{$\pm$ 0.60}} & {63.39\tiny{$\pm$ 14.12}} & {48.05\tiny{$\pm$ 14.10}} & {3.83\tiny{$\pm$ 3.23}} & {1.11\tiny{$\pm$ 1.04}} & {66.75\tiny{$\pm$ 12.49}} & {51.40\tiny{$\pm$ 12.92}} & {3.73\tiny{$\pm$ 2.81}} & {1.01\tiny{$\pm$ 0.88}} \\ \cmidrule(lr){1-1} \cmidrule(lr){3-14}
MT\tiny{(NeurIPS’17)~\cite{tarvainen2017mean}}  &  & 78.30\tiny{$\pm$6.73} & 64.83\tiny{$\pm$8.88} & 4.42\tiny{$\pm$1.96} & 1.14\tiny{$\pm$0.55} & 69.02\tiny{$\pm$14.38} & 54.46\tiny{$\pm$16.11} & 6.20\tiny{$\pm$7.58} & 1.77\tiny{$\pm$1.84} & 73.66\tiny{$\pm$12.16} & 59.65\tiny{$\pm$14.04} & 5.31\tiny{$\pm$5.61} & 1.45\tiny{$\pm$1.40}  \\
UA-MT\tiny{(MICCAI'19)~\cite{yu2019uncertainty}} &  & 78.67\tiny{$\pm$8.21} & 65.57\tiny{$\pm$10.78} & 4.55\tiny{$\pm$2.17} & \underline{1.05\tiny{$\pm$0.62}} & 69.62\tiny{$\pm$14.54} & 55.12\tiny{$\pm$15.54} & 4.79\tiny{$\pm$3.74} & 1.23\tiny{$\pm$0.94} & 74.15\tiny{$\pm$12.65} & 60.34\tiny{$\pm$14.36} & \underline{4.67\tiny{$\pm$3.06}} & \underline{1.14\tiny{$\pm$0.80}} \\
MC-Net+\tiny{(MIA'22)~\cite{wu2022mutual}}   & & 69.42\tiny{$\pm$11.24} & 54.58\tiny{$\pm$12.49} & 9.14\tiny{$\pm$5.59} & 1.97\tiny{$\pm$1.80} & 54.02\tiny{$\pm$22.44} & 40.63\tiny{$\pm$19.77} & 11.19\tiny{$\pm$8.58} & 3.05\tiny{$\pm$3.18} & 61.72\tiny{$\pm$19.69} & 47.60\tiny{$\pm$18.34} & 10.17\tiny{$\pm$7.56} & 2.51\tiny{$\pm$2.73}  \\
FUSSNet\tiny{(MICCAI'22)~\cite{xiang2022fussnet}}   & & \underline{81.97\tiny{$\pm$8.31}} & \underline{70.34\tiny{$\pm$11.60}} & \underline{4.02\tiny{$\pm$1.84}} & 1.07\tiny{$\pm$0.60} & \underline{70.21\tiny{$\pm$18.04}} & \underline{56.93\tiny{$\pm$19.73}} & 5.59\tiny{$\pm$6.27} & 1.52\tiny{$\pm$1.50} & \underline{76.09\tiny{$\pm$15.27}} & \underline{63.63\tiny{$\pm$17.57}} & 4.81\tiny{$\pm$4.77} & 1.30\tiny{$\pm$1.19} \\
CauSSL\tiny{(ICCV'23)~\cite{miao2023caussl}}   & &  53.26\tiny{$\pm$14.22} & 38.36\tiny{$\pm$12.61} & 7.85\tiny{$\pm$7.14} & 2.27\tiny{$\pm$1.08} & 43.64\tiny{$\pm$20.55} & 30.64\tiny{$\pm$16.36} & 13.83\tiny{$\pm$12.13} & 3.63\tiny{$\pm$2.35} & 48.45\tiny{$\pm$18.82} & 34.50\tiny{$\pm$15.44} & 10.84\tiny{$\pm$11.44} & 2.95\tiny{$\pm$2.16} \\
AC-MT\tiny{(MIA'23)~\cite{xu2023ambiguity}}  &  & 74.14\tiny{$\pm$8.03} & 59.71\tiny{$\pm$9.95} & 5.03\tiny{$\pm$2.35} & 1.39\tiny{$\pm$0.63} & 58.01\tiny{$\pm$13.11} & 44.56\tiny{$\pm$13.34} & 5.74\tiny{$\pm$4.83} & 1.51\tiny{$\pm$1.26} & 66.07\tiny{$\pm$14.94} & 52.13\tiny{$\pm$15.33} & 5.38\tiny{$\pm$3.95} & 1.45\tiny{$\pm$1.03}   \\
BCP\tiny{(CVPR'23)~\cite{bai2023bidirectional}} &  & 76.03\tiny{$\pm$8.39} & 62.34\tiny{$\pm$10.63} & 5.10\tiny{$\pm$3.28} & 1.56\tiny{$\pm$0.87} & 55.75\tiny{$\pm$13.93} & 44.82\tiny{$\pm$14.94} & \underline{4.68\tiny{$\pm$4.89}} & \underline{1.20\tiny{$\pm$1.30}} & 65.89\tiny{$\pm$18.29} & 53.58\tiny{$\pm$18.07} & 4.89\tiny{$\pm$4.85} & 1.38\tiny{$\pm$1.31}  \\
Su et al.\tiny{(MIA'24)~\cite{su2024mutual}} & & 72.54\tiny{$\pm$9.05} & 57.79\tiny{$\pm$10.80} & 6.76\tiny{$\pm$5.17} & 2.17\tiny{$\pm$1.50} & 62.37\tiny{$\pm$18.20} & 48.01\tiny{$\pm$17.96} & 6.52\tiny{$\pm$6.11} & 1.93\tiny{$\pm$1.79} & 67.45\tiny{$\pm$15.34} & 52.90\tiny{$\pm$15.70} & 6.64\tiny{$\pm$5.84} & 2.05\tiny{$\pm$1.69} \\ \cmidrule(lr){1-1} \cmidrule(lr){3-14}
SemiSAM+\tiny{(arXiv'25)~\cite{zhang2025semisam+}} & & 79.33\tiny{$\pm$7.70} & 66.40\tiny{$\pm$10.34} & 4.92\tiny{$\pm$5.02} & 1.26\tiny{$\pm$1.02} & 68.80\tiny{$\pm$18.22} & 55.10\tiny{$\pm$19.31} & 4.77\tiny{$\pm$5.22} & 1.38\tiny{$\pm$1.48} & 74.06\tiny{$\pm$14.95} & 60.75\tiny{$\pm$16.49} & 4.84\tiny{$\pm$5.12} & 1.32\tiny{$\pm$1.27} \\
\cellcolor{lightblue!50}\textbf{UnCoL(ours)}  & & 
\cellcolor{lightblue!50}\textbf{83.46\tiny{$\pm$8.29}} & \cellcolor{lightblue!50}\textbf{72.45\tiny{$\pm$11.84}} & \cellcolor{lightblue!50}\textbf{3.80\tiny{$\pm$2.55}} & \cellcolor{lightblue!50}\textbf{0.88\tiny{$\pm$0.64}} & \cellcolor{lightblue!50}\textbf{74.31\tiny{$\pm$17.17}} & \cellcolor{lightblue!50}\textbf{61.82\tiny{$\pm$19.79}} & \cellcolor{lightblue!50}\textbf{3.61\tiny{$\pm$2.83}} & \cellcolor{lightblue!50}\textbf{1.14\tiny{$\pm$1.02}} & \cellcolor{lightblue!50}\textbf{78.89\tiny{$\pm$14.24}} & \cellcolor{lightblue!50}\textbf{67.13\tiny{$\pm$17.16}} & \cellcolor{lightblue!50}\textbf{3.70\tiny{$\pm$2.70}} & \cellcolor{lightblue!50}\textbf{1.01\tiny{$\pm$0.86}} \\ \midrule\midrule
VNet\tiny{(3DV'16)~\cite{milletari2016v}}  & \multirow{3}{*}{\makecell{{70/70}\\{($100\%$)}}} & 76.02\tiny{$\pm$15.89} & 64.54\tiny{$\pm$17.85} & 10.05\tiny{$\pm$11.92} & 3.53\tiny{$\pm$4.85} & 68.22\tiny{$\pm$21.97} & 56.59\tiny{$\pm$23.06} & 10.80\tiny{$\pm$12.43} & 4.00\tiny{$\pm$5.12} & 72.12\tiny{$\pm$19.62} & 60.56\tiny{$\pm$21.05} & 10.43\tiny{$\pm$12.20} & 3.76\tiny{$\pm$5.02} \\
{nnU-Net\tiny{(Nat. Methods'21)~\cite{isensee2021nnu}}} &  & {85.63\tiny{$\pm$ 5.89}} & {75.43\tiny{$\pm$ 8.83}} & {3.63\tiny{$\pm$ 2.33}} & {0.69\tiny{$\pm$ 0.37}} & {76.41\tiny{$\pm$ 14.72}} & {64.22\tiny{$\pm$ 17.00}} & {3.82\tiny{$\pm$ 3.02}} & {1.23\tiny{$\pm$ 1.03}} & {81.02\tiny{$\pm$ 12.33}} & {69.83\tiny{$\pm$ 14.96}} & {3.72\tiny{$\pm$ 2.98}} & {0.96\tiny{$\pm$ 0.83}} \\
{\scriptsize{Swin-UNETR}\tiny{(MICCAI'21)~\cite{hatamizadeh2021swin}}} &  & {82.64\tiny{$\pm$ 7.85}} & {71.42\tiny{$\pm$ 10.83}} & {2.30\tiny{$\pm$ 1.46}} & {0.50\tiny{$\pm$ 0.36}} & {73.68\tiny{$\pm$ 15.51}} & {61.02\tiny{$\pm$ 16.95}} & {2.91\tiny{$\pm$ 3.12}} & {0.79\tiny{$\pm$ 0.93}} & {78.16\tiny{$\pm$ 13.34}} & {66.22\tiny{$\pm$ 15.52}} & {2.60\tiny{$\pm$ 2.57}} & {0.64\tiny{$\pm$ 0.74}} \\
\bottomrule[1.3pt]
\end{tabular}}
\end{table*}

\subsubsection{Datasets}

We evaluate UnCoL on three publicly available medical image segmentation benchmarks {that are not included in the pretraining datasets of the foundation models.}
\textbf{(1) OASIS (2D Brain MRIs)}~\cite{marcus2007open}: This dataset comprises T1-weighted brain MRIs from 414 subjects. We perform five independent trials with a 7:1:2 split for training, validation, and test, respectively. The task is four-class tissue segmentation: Left/Right Thalamus (LT, RT), Hippocampus (LH, RH). All slices are resized to $256 \times 256$ with an in-plane spacing of 1.0 $mm$ and augmented with random rotation and flipping, following~\cite{luo2022semi,bai2023bidirectional}. \textbf{(2) Pancreas-CT (3D Abdominal CT Images)}~\cite{roth2015deeporgan}: 
{This dataset contains 82 abdominal CT volumes with manual pancreas annotations. We follow the widely adopted semi-supervised setting in~\cite{wu2022mutual,xiang2022fussnet,lu2023upcol}, using 62 volumes for training and 20 for testing, with the labeled subset identical to that used in prior works to ensure fair comparison.}
\textbf{(3) ImageTBAD (3D Aortic CTA Scans)}~\cite{yao2021imagetbad}: This dataset contains 100 thoracic CTA scans annotated for type B aortic dissection. 
{We follow same 7:1:2 train/validation/test split at the volume level across five trials.}
To ensure compatibility with the input size of the foundation models, all volumes are first resampled to an isotropic resolution of 1.0 $mm^3$ and then uniformly resized to a fixed shape of $128^3$ voxels using trilinear interpolation. {Results are reported as mean and standard deviation across trials.}

\subsubsection{Model Setup}
We employ MedSAM~\cite{ma2024segment} and SAM-Med3D~\cite{wang2023sam} as the frozen 2D and 3D foundation models, respectively. Both models use a 12-layer Vision Transformer (ViT) as the image encoder, with a hidden dimension of 768. The student model employs a lightweight 4-layer SimpleViT encoder, with a task-specific backbone: U-Net~\cite{ronneberger2015u} for 2D and V-Net~\cite{milletari2016v} for 3D segmentation. Cross-layer distillation is conducted between the 4-layer transformer encoder of the student model and the 12-layer image encoder of the generalized teacher. 
Specifically, the first three student layers are aligned with the 4th, 8th, and 12th layers of the teacher encoder, {to capture progressively higher levels of semantic priors.} The final student layer is supervised by the output of the teacher’s image-prompt fusion module.

\subsubsection{Training Details}
\label{sec:exp_set_td}
All models are implemented in PyTorch and trained on a single NVIDIA RTX 3090 GPU. Training uses SGD with a learning rate of 0.01 and a weight decay of $10^{-4}$, for 15{,}000 iterations in pretraining and fine-tuning stages. During pretraining (Stage~1), each training batch consists of 4 labeled samples, while in fine-tuning (Stage~2), each batch includes 2 labeled and 2 unlabeled samples. 
Throughout training, synthetic prompts are {supplied to} the frozen generalized teacher to guide supervision, 
while the student model performs prompt-free inference.
For 2D inputs, bounding-box prompts are created by shifting ground-truth annotations or pseudo-label masks by 20 pixels. For 3D inputs, point prompts are sampled either from labeled voxels or from regions with low predictive uncertainty.
Following~\cite{tarvainen2017mean}, we adopt a Gaussian ramp-up function $\gamma(t) = \exp(-5(1 - t/t_{\text{max}})^2)$, where $t$ denotes the current iteration and $t_{\text{max}} = 15{,}000$, to progressively increase the supervision strength. The distillation weight is set as $\alpha(t) = \lambda_{\text{vis}} \cdot \gamma(t)$ with $ \lambda_{\text{vis}}=0.1$, while the entropy threshold is defined as $\tau(t) = 0.75 + 0.25 \cdot \gamma(t)$ to gradually relax the filtering criterion~\cite{yu2019uncertainty}. Only voxels with entropy below $\tau(t)$ are supervised, with a fixed loss weight $\lambda_{\text{pseudo}} = 0.5$~\cite{bai2023bidirectional}.

\subsubsection{Evaluation Protocols}
Model performance is assessed using Dice Similarity Coefficient (DSC), Jaccard Index, 95\% Hausdorff Distance (95HD), and Average Symmetric Surface Distance (ASD).
We compare UnCoL against five categories of baselines: 
\textbf{(1) Foundation Models:} SAM~\cite{kirillov2023segment}, MedSAM~\cite{ma2024segment}, SAM-Med3D~\cite{wang2023sam}, {SegVol~\cite{du2024segvol}}, {using the official prompting protocol of each model}.
\textbf{(2) Classical SSL:} Mean Teacher (MT)~\cite{tarvainen2017mean} and UAMT~\cite{yu2019uncertainty}.
\textbf{(3) Recent SSL:} MC-Net+~\cite{wu2022mutual}, FUSSNet~\cite{xiang2022fussnet}, CassuSSL~\cite{miao2023caussl}, AC-MT~\cite{xu2023ambiguity}, URPC~\cite{luo2022semi}, BCP~\cite{bai2023bidirectional}, ABD~\cite{chi2024adaptive}, and Su et al.~\cite{su2024mutual}.
\textbf{(4) SAM-Enhanced SSL:} CPC-SAM~\cite{miao2024cross} (2D only) and SemiSAM+~\cite{zhang2025semisam+} (3D only).
{\textbf{(5) Supervised Baselines:} VNet~\cite{milletari2016v}, nnU-Net~\cite{isensee2021nnu}, and Swin-UNETR~\cite{hatamizadeh2021swin}}.
All baselines are trained and evaluated under identical data splits and protocols. Official implementations are used {when} available, while methods lacking native multi-class support are accordingly adapted based on their {source} code.
{Swin-UNETR~\cite{hatamizadeh2021swin} and SegVol~\cite{du2024segvol} are volumetric architectures tailored for 3D segmentation and are evaluated on Pancreas-CT and ImageTBAD, whereas 2D methods are compared on OASIS.}

\subsection{Quantitative Results}\label{sec:quantresults}
\subsubsection{Comparison on 2D MRI Segmentation (OASIS)}
We evaluate UnCoL on the OASIS dataset across different annotation ratios. As shown in Table~\ref{tab:oasis}, UnCoL consistently outperforms existing methods using only 5-10\% labeled data. 
While MedSAM~\cite{ma2024segment}, pretrained on brain MRI datasets like FeTA~\cite{dorent2023crossmoda}, achieves reasonable zero-shot performance (average Dice score of 76.49\%), its effectiveness is limited by anatomical and distributional shifts{, reflecting the general limitations of zero-shot foundation models under domain shift. UnCoL overcomes this by using uncertainty-guided supervision to adapt effectively with minimal labels.} 

With only 5\% labeled data, UnCoL achieves an average Dice score of 94.07\%, surpassing ABD~\cite{chi2024adaptive} (92.01\%) and even the fully supervised baseline (92.20\%). It also reduces 95HD from 4.17 to 2.18 voxel units and maintains low ASD and standard deviation.
{CPC-SAM~\cite{miao2024cross}, a recent SAM-based SSL method that also relies on MedSAM, attains 91.33\% Dice on 2D datasets. Compared with its prompt-driven consistency design, UnCoL delivers higher accuracy with a simpler and prompt-free formulation, avoiding prompt sensitivity and architectural overhead.}

{Figure~\ref{fig:tbad} further demonstrates improvements in challenging regions, particularly in symmetric structures such as the left/right thalamus and hippocampus, where conventional SSL approaches often produce inconsistent boundaries.}

\subsubsection{3D CT Segmentation (Pancreas-CT and ImageTBAD)}
We further evaluate UnCoL on two challenging 3D segmentation benchmarks. Notably, UnCoL and SemiSAM+~\cite{zhang2025semisam+} are the only SAM-enhanced semi-supervised segmentation methods applied across both datasets, utilizing SAM-Med3D as the foundational model. Table~\ref{tab:pancreas} presents results on the Pancreas-CT dataset. With 20\% labeled data, UnCoL achieves a DSC of 83.16\% and an ASD of 1.44 voxels, outperforming the fully supervised baseline, which achieves 81.89\% DSC and 2.39 voxels ASD. With only 10\% labeled data, UnCoL attains 80.40\% Dice, along with the lowest 95HD of 6.86 voxels and ASD of 1.85 voxels. 
{Despite the pancreas being included in the pretraining corpus of SAM-Med3D (e.g., AbdomenCT-1K~\cite{ma2021abdomenct}), dataset-level differences still hinder zero-shot performance.} UnCoL addresses this limitation through uncertainty-informed dual-teacher collaboration, enabling anatomically consistent segmentation even with limited supervision.

{On ImageTBAD, we further perform a class-wise analysis on the two clinically critical structures, True Lumen (TL) and False Lumen (FL). As shown in Table~\ref{tab:tbad}, TL is generally easier to segment due to its larger and more stable morphology, whereas FL exhibits substantial shape variability and weaker contrast.
With 20\% labels, UnCoL improves TL Dice from 70.12\% to 83.46\% (+13.34\%) and FL Dice from 51.95\% to 74.31\% (+22.36\%), indicating that our framework provides the largest gains on the most challenging class (FL).}

{We observe two major failure modes in prior methods in Fig.~\ref{fig:tbad}: (1) TL-FL boundary leakage in the dissection plane due to low contrast, and (2) misclassification of thrombosed FL as background or TL, especially near the entry tear. 
{Clinically, accurate TL-FL separation is essential for assessing blood flow, lumen geometry, and dissection subtype, while FL extent and morphology are closely linked to disease severity and prognosis~\cite{evangelista2018insights}. TL-FL delineation errors are therefore among the most critical, as they may distort lumen measurements and obscure high-risk regions. UnCoL reduces boundary leakage and better preserves low-contrast, heterogeneous FL regions by integrating global vascular priors with localized refinements, resulting in more clinically consistent TL-FL separation, particularly in high-uncertainty FL areas.}

\begin{table*}[htbp]
\centering
\caption{Ablation study on the 2D and 3D datasets with $10\%$ labeled data.  
95HD and ASD are in voxel units.
{Abbreviations: Stu. - Student, G-Tch - Generalized teacher, S-Tch - Specialized teacher, DPKD - Dual-Path Knowledge Distillation, UAPL - Uncertainty-Aware Pseudo-Labeling.
$\mathcal{L}_{\text{vis}}$ and $\mathcal{L}_{\text{sem}}$ denote the visual and semantic distillation losses.
$\mathcal{L}_{\text{pseudo}}$ compares different pseudo-label supervision strategies, where S-Tch-PL uses pseudo-labels generated by the specialized teacher, Avg-Tch-PL averages the predictions of both teachers, and UAPL selects confident and consistent pseudo-labels via uncertainty guidance.}
} \label{tab:abl}
\resizebox{0.98\textwidth}{!}{
\begin{tabular}{l|p{0.6cm}cc|p{0.6cm}cc|llll|llll|llll}
\toprule[1.3pt]
\multirow{2}{*}{Method}  & \multicolumn{3}{c}{Model} & 
\multicolumn{2}{|c}{DPKD} & UAPL
& \multicolumn{4}{c|}{OASIS} & \multicolumn{4}{c|}{Pancreas-CT}  & \multicolumn{4}{c}{ImageTBAD}  \\ \cmidrule{2-19}

 & \multicolumn{1}{c}{Stu.} & \multicolumn{1}{c}{G-Tch}  & \multicolumn{1}{c|}{S-Tch}  & \multicolumn{1}{c}{$\mathcal{L}_{\text{vis}}$} & $\mathcal{L}_{\text{sem}}$  & $\mathcal{L}_{\text{pseudo}}$  & DSC$(\%)\uparrow$ & \scriptsize{Jaccard$(\%)\uparrow$} & 95HD$\downarrow$   & ASD$\downarrow$  & DSC$(\%)\uparrow$ & \scriptsize{Jaccard$(\%)\uparrow$} & 95HD$\downarrow$   & ASD$\downarrow$  & DSC$(\%)\uparrow$ & \scriptsize{Jaccard$(\%)\uparrow$} & 95HD$\downarrow$   & ASD$\downarrow$  \\ \midrule

\multirow{3}{*}{Baseline} &\multicolumn{1}{c}{\checkmark} &  &   & &  & & 66.39\tiny{$\pm$5.55} & 56.71\tiny{$\pm$5.04} & 24.01\tiny{$\pm$1.55} & 9.84\tiny{$\pm$3.17}  & 56.96\tiny{$\pm$16.50} & 41.38\tiny{$\pm$13.66} & 18.69\tiny{$\pm$12.60} & 5.38\tiny{$\pm$3.90} & 55.85\tiny{$\pm$21.20} & 42.41\tiny{$\pm$18.96} & 14.19\tiny{$\pm$11.74} & 3.65\tiny{$\pm$4.56} \\ 

 &  & \multicolumn{1}{c}{\checkmark} &   & &  & & 76.49\tiny{$\pm$0.67} & 62.68\tiny{$\pm$0.65} & 7.30\tiny{$\pm$0.07} & 2.23\tiny{$\pm$0.04} & \underline{79.37\tiny{$\pm$}4.55} & \underline{66.02\tiny{$\pm$}6.22} & \textbf{5.52\tiny{$\pm$}1.79} & \textbf{1.55\tiny{$\pm$}0.54} & 29.61\tiny{$\pm$}19.11 & 19.04\tiny{$\pm$}14.43 & 13.23\tiny{$\pm$}5.58 & 4.44\tiny{$\pm$}1.94  \\
 
 &\multicolumn{1}{c}{\checkmark} &  & \multicolumn{1}{c|}{\checkmark}  & &  & & 84.02\tiny{$\pm$4.15} & 76.41\tiny{$\pm$3.80} & 12.86\tiny{$\pm$3.58} & 4.37\tiny{$\pm$1.72} & 64.28\tiny{$\pm$11.62} & 48.40\tiny{$\pm$12.10} & 28.19\tiny{$\pm$14.54} & 8.92\tiny{$\pm$5.15} & 64.67\tiny{$\pm$15.93} & 50.04\tiny{$\pm$15.83} & 7.48\tiny{$\pm$6.06} & 2.00\tiny{$\pm$1.43}  \\ \midrule

\multirow{3}{*}{\makecell{Data-Efficient\\Pretraining}}  &\multicolumn{1}{c}{\checkmark} & \multicolumn{1}{c}{\checkmark}  &   & \multicolumn{1}{c}{\checkmark} &  & & 93.65\tiny{$\pm$2.54} & 88.16\tiny{$\pm$4.26} & 3.14\tiny{$\pm$9.13} & 0.90\tiny{$\pm$1.62} &73.22\tiny{$\pm$9.95} & 58.68\tiny{$\pm$11.44} & 10.39\tiny{$\pm$6.57} & 3.38\tiny{$\pm$1.52} & 52.83\tiny{$\pm$16.38} & 41.27\tiny{$\pm$16.88} & \textbf{4.43\tiny{$\pm$4.75}} & \textbf{1.30\tiny{$\pm$1.27}}   \\

  &\multicolumn{1}{c}{\checkmark} & \multicolumn{1}{c}{\checkmark}  &   &  & \multicolumn{1}{c}{\checkmark} & & 93.17\tiny{$\pm$2.17} & 87.29\tiny{$\pm$3.71} & 2.39\tiny{$\pm$5.52} & 0.83\tiny{$\pm$1.28}  &72.66\tiny{$\pm$9.96} & 57.98\tiny{$\pm$11.68} & 11.90\tiny{$\pm$5.92} & 3.76\tiny{$\pm$1.60}  & 67.63\tiny{$\pm$16.48} & 53.33\tiny{$\pm$17.06} & 5.44\tiny{$\pm$4.84} & 1.63\tiny{$\pm$1.60}  \\
  
  &\multicolumn{1}{c}{\checkmark} & \multicolumn{1}{c}{\checkmark}  &   &\multicolumn{1}{c}{\checkmark}  & \multicolumn{1}{c}{\checkmark} &  & \underline{93.69\tiny{$\pm$3.75}} & \underline{88.34\tiny{$\pm$6.03}} & \textbf{1.95\tiny{$\pm$0.94}} & \textbf{0.61\tiny{$\pm$0.50}}  &74.12\tiny{$\pm$9.49} & 59.74\tiny{$\pm$11.09} & 10.79\tiny{$\pm$6.63} & 3.38\tiny{$\pm$1.78}  & 69.73\tiny{$\pm$17.56} & 56.02\tiny{$\pm$18.65} & 5.92\tiny{$\pm$7.43} & 1.64\tiny{$\pm$1.38}  \\ \midrule
  
\multirow{3}{*}{\makecell{Semi-Supervised\\Fine-tuning}} &\multicolumn{1}{c}{\checkmark} & \multicolumn{1}{c}{\checkmark}  &  \multicolumn{1}{c|}{\checkmark} &\multicolumn{1}{c}{\checkmark}  & \multicolumn{1}{c}{\checkmark} & \scriptsize{{S-Tch-PL}} & {93.08\tiny{$\pm$2.32}} & {87.14\tiny{$\pm$3.91}} & {2.44\tiny{$\pm$5.28}} & {0.77\tiny{$\pm$0.87}} & {80.08\tiny{$\pm$6.11}} & {67.18\tiny{$\pm$8.05}} & {7.20\tiny{$\pm$4.10}} & {2.15\tiny{$\pm$1.08}} & {\textbf{71.26\tiny{$\pm$1.58}}} & {\textbf{57.44\tiny{$\pm$1.54}}} & {4.84\tiny{$\pm$1.75}} & {1.38\tiny{$\pm$0.40}}  \\
&\multicolumn{1}{c}{\checkmark} & \multicolumn{1}{c}{\checkmark}  &  \multicolumn{1}{c|}{\checkmark} &\multicolumn{1}{c}{\checkmark}  & \multicolumn{1}{c}{\checkmark} & \scriptsize{{Avg-Tch-PL}} & {92.97\tiny{$\pm$2.32}} & {86.95\tiny{$\pm$3.90}} & {2.48\tiny{$\pm$5.24}} & {0.76\tiny{$\pm$0.77}} & {77.29\tiny{$\pm$7.58}} & {63.56\tiny{$\pm$9.20}} & {7.57\tiny{$\pm$5.13}} & {2.32\tiny{$\pm$1.40}} & {70.79\tiny{$\pm$1.68}} & {57.16\tiny{$\pm$1.22}} & {4.87\tiny{$\pm$1.77}} & {\textbf{1.30\tiny{$\pm$0.40}}}  \\
&\multicolumn{1}{c}{\checkmark} & \multicolumn{1}{c}{\checkmark}  &  \multicolumn{1}{c|}{\checkmark} &\multicolumn{1}{c}{\checkmark}  & \multicolumn{1}{c}{\checkmark} & \scriptsize{UAPL} & \cellcolor{lightblue!50}\textbf{94.42\tiny{$\pm$0.29}} & \cellcolor{lightblue!50}\textbf{89.51\tiny{$\pm$0.47}} & \cellcolor{lightblue!50}\underline{2.06\tiny{$\pm$3.27}} & \cellcolor{lightblue!50}\textbf{0.61\tiny{$\pm$0.63}}  & \cellcolor{lightblue!50}\textbf{80.40\tiny{$\pm$5.82}} & \cellcolor{lightblue!50}\textbf{67.60\tiny{$\pm$7.83}} & \cellcolor{lightblue!50}\underline{6.86\tiny{$\pm$3.96}} & \cellcolor{lightblue!50}\underline{1.85\tiny{$\pm$0.94}} & \cellcolor{lightblue!50}\underline{70.97\tiny{$\pm$6.35}} & \cellcolor{lightblue!50}\underline{57.17\tiny{$\pm$6.81}} & \cellcolor{lightblue!50}\underline{4.83\tiny{$\pm$2.22}} & \cellcolor{lightblue!50}\textbf{1.30\tiny{$\pm$0.58}}  \\
\bottomrule[1.3pt]
\end{tabular}}
\end{table*}

\subsection{Ablation Study}
\subsubsection{{Effect of Teacher Model Configuration}}
{To assess the influence of different teacher configurations, we first compare the baseline model without teacher guidance, single-teacher settings, and the dual-teacher setup.}
As shown in the top block of Table~\ref{tab:abl}, the Generalized Teacher (G{-Tch}) increases Dice {over the student-only baseline} from 66.39\% to 76.49\% on OASIS and from 56.96\% to 79.37\% on Pancreas-CT{, demonstrating effective prior transfer from the foundation model}. However, {when encountering unseen anatomy, }{its performance collapses to 29.61\% on ImageTBAD (vs. 55.85\% for the student), even with ground-truth-derived prompts, suggesting its domain dependency}. 
The Specialized Teacher (S{-Tch}), derived via an EMA of the student model, {provides stable gains (84.02\%, 64.28\%, and 64.67\%) but underperforms the generalized teacher on well-studied domains (e.g., 64.28\% vs. 79.37\% on Pancreas-CT).}
{These results show that neither teacher alone is consistently reliable, motivating a dual-teacher design to balance generalization and specialization. Since directly combining two imperfect teachers can propagate errors, DPKD and UAPL are introduced to selectively transfer reliable knowledge, as validated in later ablations.
}

\subsubsection{{Effectiveness of Dual-Path Knowledge Distillation}}
The middle block {of Table~\ref{tab:abl}} evaluates the impact of {our Dual-Path Knowledge Distillation (DPKD)} during the data-efficient pretraining stage. {This stage transfers generalized knowledge from the foundation model through two complementary paths: visual distillation and semantic distillation.} Visual distillation alone yields Dice scores of 93.65\% on OASIS, 73.22\% on Pancreas-CT, and 52.83\% on ImageTBAD. 
{Semantic distillation alone, which aligns class-level semantics using prompt-guided features, also yields strong improvements—93.17\%, 72.66\%, and 67.63\%.
When combining both paths, the Dice scores reach 93.69\%, 74.12\%, and 69.73\%, corresponding to overall gains of +27.30\%, +17.16\%, and +13.88\% over the student-only baseline, respectively.}
{This dual-path design jointly captures visual and contextual cues, enabling prompt-free inference in the subsequent fine-tuning stage and effectively bridging the gap between foundation-level generalization and dataset-specific specialization.}

\begin{table}[htbp]
\centering
\caption{{Evaluation of pseudo-label generation quality after data-efficient pretraining
using 10\% labeled data. 
Pseudo-label Sources: G-Tch - Generalized Teacher, S-Tch - Specialized Teacher, Stu - Student.}}
\label{tab:plquality}
\resizebox{0.49\textwidth}{!}{
\begin{tabular}{l l c c c c}
\toprule[1.3pt]
Dataset & Pseudo-label Source & DSC(\%) $\uparrow$ & Jaccard(\%) $\uparrow$ & 95HD $\downarrow$ & ASD $\downarrow$ \\
\midrule
\multirow{3}{*}{OASIS (2D)}
& G-Tch            & 82.11\tiny{$\pm$6.62}	&	70.17\tiny{$\pm$9.05}	&	4.57\tiny{$\pm$1.34}	&	1.74\tiny{$\pm$0.50}  \\
& S-Tch       & {94.35\tiny{$\pm$2.26}}	&	{89.40\tiny{$\pm$3.89}}	&	{1.95\tiny{$\pm$3.75}}	&	{0.61\tiny{$\pm$0.85}} \\
& Stu         & {93.69\tiny{$\pm$3.75}} & {88.34\tiny{$\pm$6.03}} & {{1.95\tiny{$\pm$0.94}}} & {{0.61\tiny{$\pm$0.50}}}  \\
\midrule
\multirow{3}{*}{Pancreas-CT (3D)}
& G-Tch            & 60.24\tiny{$\pm$20.88} & 45.98\tiny{$\pm$19.44} & 22.90\tiny{$\pm$17.10} & 9.00\tiny{$\pm$8.55} \\
& S-Tch       & 72.77\tiny{$\pm$10.34} & 58.16\tiny{$\pm$11.91} & 9.00\tiny{$\pm$4.86} & 2.67\tiny{$\pm$1.45}  \\
& Stu         & {74.12\tiny{$\pm$9.49}} & {59.74\tiny{$\pm$11.09}} & 10.79\tiny{$\pm$6.63} & 3.38\tiny{$\pm$1.78}  \\
\midrule
\multirow{3}{*}{ImageTBAD (3D)}
& G-Tch            & 15.98\tiny{$\pm$14.01} & 9.38\tiny{$\pm$8.93} & 21.83\tiny{$\pm$7.62} & 8.70\tiny{$\pm$4.47}  \\
& S-Tch       &  66.48\tiny{$\pm$17.18} & 52.28\tiny{$\pm$17.41} & {5.86\tiny{$\pm$6.02}} & {1.70\tiny{$\pm$1.76}}  \\
& Stu         & {69.73\tiny{$\pm$17.56}} & {56.02\tiny{$\pm$18.65}} & 5.92\tiny{$\pm$7.43} & 1.64\tiny{$\pm$1.38}  \\
\bottomrule[1.3pt]
\end{tabular}}
\end{table}

\subsubsection{{Pseudo-Label Quality Analysis}}
{To assess the reliability of teacher predictions after data-efficient pretraining, we evaluate the pseudo-label quality of the generalized teacher (G-Tch), specialized teacher (S-Tch), and student (Stu) before semi-supervised fine-tuning. 
The S-Tch first predicts pseudo-labels, from which prompt cues such as low-uncertainty points in 3D or expanded boxes in 2D are derived to guide the G-Tch in generating prompt-free predictions. 
This process ensures consistency with the fine-tuning stage.}

{As shown in Table~\ref{tab:plquality}, the S-Tch produces more accurate and stable pseudo-labels than the G-Tch, achieving Dice scores of 94.35\%, 72.77\%, and 66.48\% on OASIS, Pancreas-CT, and ImageTBAD, respectively.
This improvement is consistent with the effect of transferring generalized semantics, as evidenced by the S-Tch baseline without G-Tch assistance in Table~\ref{tab:abl} (+10.33\%, +8.49\%, and +1.61\% Dice gains). 
The G-Tch performs well on OASIS (82.11\%) and Pancreas-CT (60.24\%) but drops to 15.98\% on ImageTBAD, indicating limited domain adaptability
{at the prediction level.
In contrast, the student improves from the specialized teacher to 69.73\% Dice on ImageTBAD (+3.25\%), showing that representation-level knowledge can still be effectively leveraged, even when G-Tch's pixel-wise predictions are inaccurate.
}
Overall, these results suggest that our pretraining yields reliable pseudo-labels that adapt well to dataset-specific characteristics, providing a solid foundation for uncertainty-aware fine-tuning.
}

\subsubsection{{Effectiveness of Uncertainty-Aware Pseudo-Labeling}}

The bottom block {of Table~\ref{tab:abl}} shows the effect of uncertainty-aware pseudo-labeling (UAPL) during fine-tuning.
{In this stage, pseudo-labels are generated from either the specialized teacher (S-Tch-PL) or the averaged outputs of both teachers (Avg-Tch-PL), but neither consistently improves over the pretrained stage. On OASIS, both variants can degrade performance, indicating that unfiltered teacher signals may introduce noise. On ImageTBAD, S-Tch-PL yields slightly higher Dice but larger boundary errors than Avg-Tch-PL, indicating that simple fusion cannot resolve teacher inconsistency.
In contrast, UAPL achieves the best and most stable results—higher Dice and lower 95HD/ASD across datasets, demonstrating that uncertainty-guided fusion improves pseudo-label reliability and geometric precision.

\begin{figure}
    \centering
    \includegraphics[width=\linewidth]{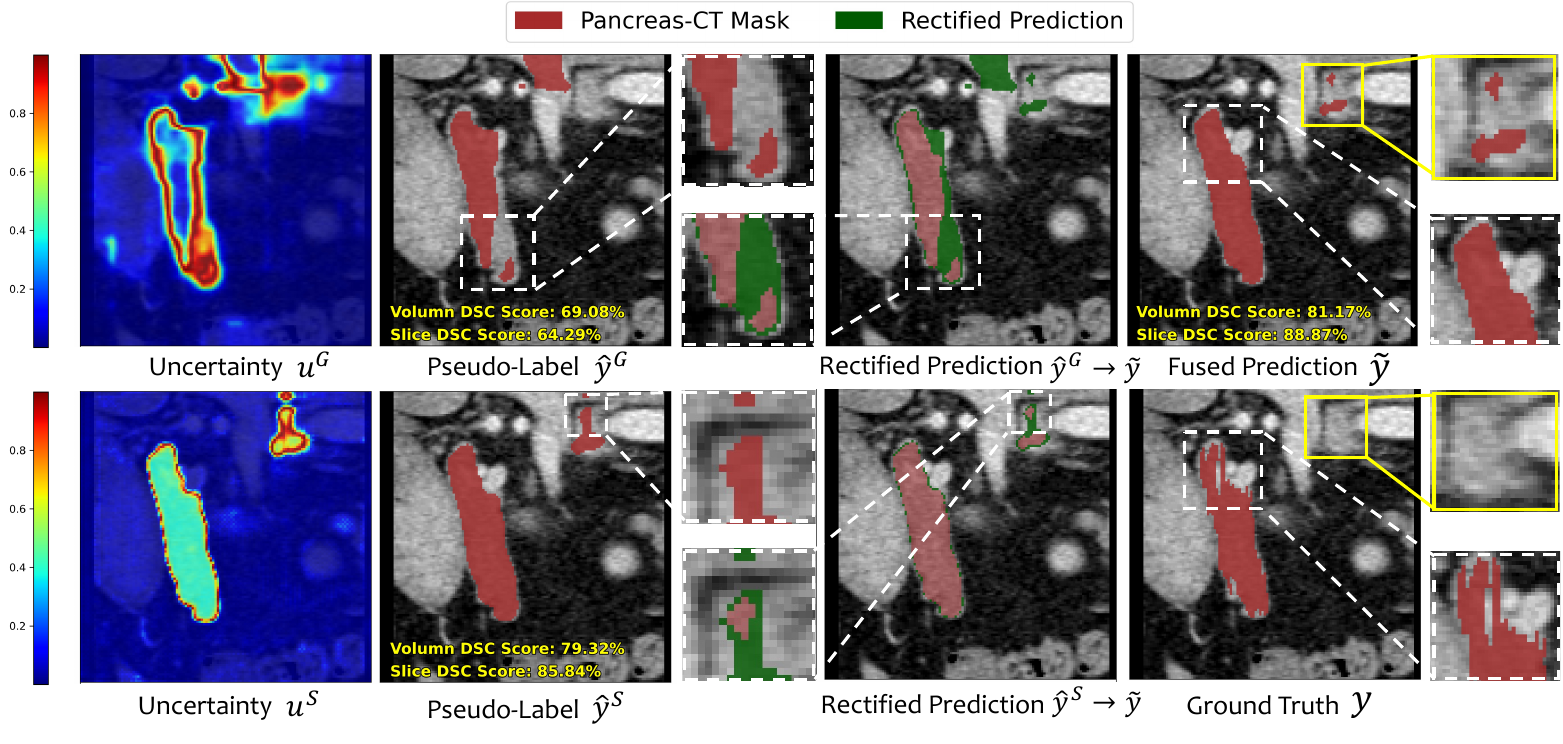}
    \caption{
    Visualization of the uncertainty-aware pseudo-label rectification process. From left to right: uncertainty maps ($u^G$, $u^S$), initial pseudo-labels ($\hat{y}^G$, $\hat{y}^S$), rectified predictions ({$\hat{y}^{G}\!\rightarrow\!\tilde{y}$, $\hat{y}^{S}\!\rightarrow\!\tilde{y}$}), the fused pseudo-label ($\tilde{y}$), and ground truth ($y$). 
    {Green regions mark changes introduced by rectification.
    White dashed boxes denote areas corrected by UAPL, and yellow solid boxes highlight residual discrepancies with respect to the ground truth.}
    }
    \label{fig:uncertain}
\end{figure}

{Figure}~\ref{fig:uncertain} illustrates this process. 
Compared with either teacher’s output, the fused pseudo-label provides not only overall improvement but also corrects localized errors for both teachers, as indicated by the white dashed boxes.
These regions correspond to areas of high uncertainty (reddish zones in the heatmaps), where UAPL adaptively replaces unreliable predictions with confident estimates from the complementary teacher.
The yellow solid boxes mark residual deviations in confident regions, showing that even strong consensus may retain slight deviation.
Compared with both teachers and the coarse ground truth, the fused prediction exhibits smoother boundaries and more connected structures, leading to more coherent anatomical structures overall.}

\begin{figure}
    \centering
    \includegraphics[width=\linewidth]{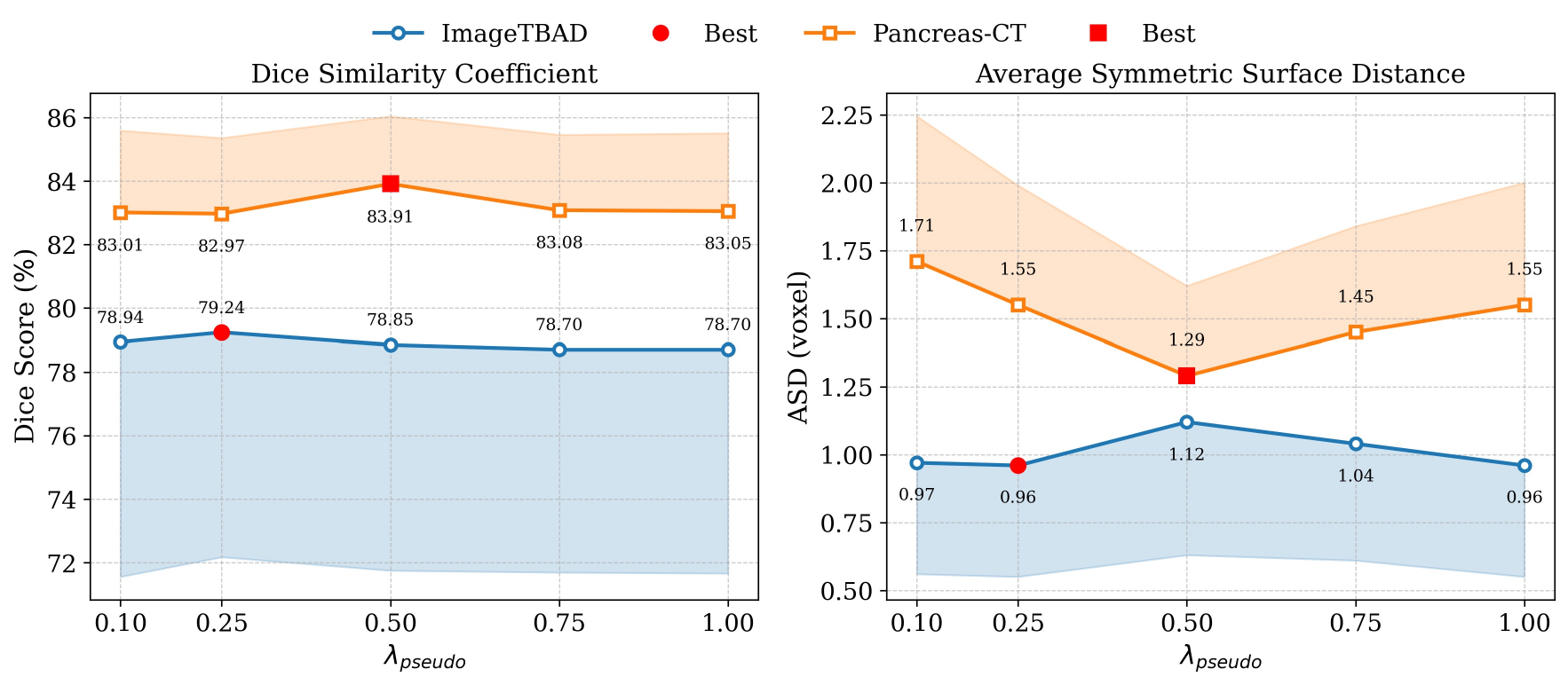}
    \caption{Sensitivity of the pseudo-label balancing coefficient $\lambda_{\text{pseudo}}$, {where performance remains stable and peaks} near 0.5.} 
    \label{fig:sensitivity}
\end{figure}

\subsubsection{Effect of \texorpdfstring{$\lambda_{\text{pseudo}}$}{lambda_pseudo} on Model Performance}

We evaluate the sensitivity of UnCoL to the pseudo-label balancing coefficient $\lambda_{\text{pseudo}}$, which controls the trade-off between supervision from ground-truth labels and pseudo-labels during semi-supervised training. Consistent with prior work~\cite{bai2023bidirectional}, we set $\lambda_{\text{pseudo}} = 0.5$ as the default.
Figure~\ref{fig:sensitivity} shows that UnCoL maintains strong performance across a wide range of $\lambda_{\text{pseudo}}$ values. On ImageTBAD, the highest Dice score of 79.24\% is obtained at $\lambda_{\text{pseudo}}=0.25$, while Pancreas-CT reaches its peak Dice of 83.91\% at $\lambda_{\text{pseudo}}=0.50$. Boundary-based metrics such as the ASD exhibit similar stability.
These results {suggest that UnCoL is robust to} the choice of $\lambda_{\text{pseudo}}$, with the uncertainty-aware pseudo-labeling mechanism consistently improving performance without the need for precise hyperparameter tuning.

\subsection{{Comparison with Classic Fine-tuning Strategies}}\label{sec:FT}
\begin{table}[htbp]
\centering
\caption{{Comparison of fine-tuning strategies on 2D/3D datasets. Our method freezes the foundation model. {0\%} and {100\%} denote zero-shot lower and fully supervised upper bounds. Metrics are averaged over all classes. Parameter sizes: Vanilla (7.58M), PEFT (296.83K), LoRA (35.33K, rank=4). Bold and \underline{underline} indicate best and second-best results.}
}
\resizebox{0.48\textwidth}{!}{
\begin{tabular}{l c l  c c c c}
\toprule
Dataset & Labeled Ratio & Method &
DSC(\%)\,$\uparrow$ & Jaccard(\%)\,$\uparrow$ & 95HD\,$\downarrow$ & ASD\,$\downarrow$ \\
\midrule
\multirow{12}{*}{OASIS} 
& {0\%} & {MedSAM} & {76.49\tiny{$\pm$0.67}} & {62.68\tiny{$\pm$0.65}} & {7.30\tiny{$\pm$0.07}} & {2.23\tiny{$\pm$0.04}} \\
\cmidrule(lr){2-7}
& \multirow{4}{*}{5\%}
& Vanilla & \pmstd{71.15}{7.35} & \pmstd{55.73}{8.51} & \pmstd{9.11}{5.36} & \pmstd{3.93}{1.20} \\
& & PEFT    & \pmstd{72.30}{7.40} & \pmstd{57.25}{8.86} & \pmstd{6.97}{1.64} & \pmstd{3.33}{0.74} \\
& & LoRA    & \underline{\pmstd{81.05}{6.46}} & \underline{\pmstd{68.67}{8.80}} & \underline{\pmstd{5.08}{3.35}} & \underline{\pmstd{2.10}{0.88}} \\
& & \textbf{UnCoL (Ours)} & \textbf{\pmstd{94.07}{2.92}} & \textbf{\pmstd{88.93}{4.78}} & \textbf{\pmstd{2.18}{4.63}} & \textbf{\pmstd{0.67}{1.28}} \\
\cmidrule(lr){2-7}
& \multirow{4}{*}{10\%}
& Vanilla & \pmstd{73.86}{7.62} & \pmstd{59.14}{9.16} & \pmstd{11.63}{8.68} & \pmstd{3.77}{1.69} \\
& & PEFT    & \pmstd{73.77}{6.32} & \pmstd{58.87}{7.61} & \pmstd{7.25}{3.93} & \pmstd{3.27}{1.02} \\
& & LoRA    & \underline{\pmstd{82.81}{5.94}} & \underline{\pmstd{71.11}{8.30}} & \underline{\pmstd{4.34}{1.25}} & \underline{\pmstd{1.76}{0.46}} \\
& & \cellcolor{lightblue!50}\textbf{UnCoL (Ours)} & \cellcolor{lightblue!50}\textbf{\pmstd{94.42}{2.17}} & \cellcolor{lightblue!50}\textbf{\pmstd{89.51}{3.77}} & \cellcolor{lightblue!50}\textbf{\pmstd{2.06}{4.07}} & \cellcolor{lightblue!50}\textbf{\pmstd{0.61}{0.87}} \\
\cmidrule(lr){2-7}
& \multirow{3}{*}{{100\%}}
& {Vanilla} & {\pmstd{79.34}{6.11}} & {\pmstd{66.20}{8.13}} & {\pmstd{5.26}{1.34}} & {\pmstd{2.35}{0.60}} \\
& & {PEFT   } & {\pmstd{79.19}{4.19}} & {\pmstd{65.75}{5.50}} & {\pmstd{4.97}{1.01}} & {\pmstd{2.35}{0.60}} \\
& & {LoRA   } & {\pmstd{85.76}{4.81}} & {\pmstd{75.37}{7.06}} & {\pmstd{3.69}{1.01}} & {\pmstd{1.43}{0.38}} \\
\midrule
\multirow{12}{*}{Pancreas-CT} 
& {0\%} & {SAM-Med3D} & {56.96\tiny{$\pm$16.50}} & {41.38\tiny{$\pm$13.66}} & {18.69\tiny{$\pm$12.60}} & {5.38\tiny{$\pm$3.90}} \\
\cmidrule(lr){2-7}
& \multirow{4}{*}{10\%}
& Vanilla & \pmstd{77.27}{5.45} & \pmstd{63.27}{7.01} & \underline{\pmstd{6.15}{2.59}} & \underline{\pmstd{1.78}{0.80}} \\
& & PEFT    & \pmstd{78.87}{4.64} & \pmstd{65.35}{6.32} & \pmstd{6.37}{4.36} & \pmstd{2.01}{1.38} \\
& & LoRA    & \underline{\pmstd{79.25}{4.84}} & \underline{\pmstd{65.89}{6.56}} & \textbf{\pmstd{5.75}{2.03} }& \textbf{\pmstd{1.72}{0.62}} \\
& & \cellcolor{lightblue!50}\textbf{UnCoL (Ours)} & \cellcolor{lightblue!50}\textbf{\pmstd{80.40}{5.82}} & \cellcolor{lightblue!50}\textbf{\pmstd{67.60}{7.83}} & \cellcolor{lightblue!50}\pmstd{6.86}{3.96} & \cellcolor{lightblue!50}\pmstd{1.85}{0.94} \\
\cmidrule(lr){2-7}
& \multirow{4}{*}{20\%}
& Vanilla & \pmstd{80.10}{4.83} & \pmstd{67.08}{6.60} & \textbf{\pmstd{5.16}{1.91}} & \underline{\pmstd{1.51}{0.54}} \\
& & PEFT    & \pmstd{79.76}{4.49} & \pmstd{66.57}{6.11} & \pmstd{5.70}{2.69} & \pmstd{1.75}{0.89} \\
& & LoRA    & \underline{\pmstd{80.65}{4.03}} & \underline{\pmstd{67.76}{5.60}} & \underline{\pmstd{5.26}{2.60}} & \pmstd{1.58}{0.73} \\
& & \cellcolor{lightblue!50}\textbf{UnCoL (Ours)} & \cellcolor{lightblue!50}\textbf{\pmstd{83.16}{4.80}} & \cellcolor{lightblue!50}\textbf{\pmstd{71.46}{6.79}} & \cellcolor{lightblue!50}\pmstd{5.44}{3.26} & \cellcolor{lightblue!50}\textbf{\pmstd{1.44}{0.90}} \\
\cmidrule(lr){2-7}
& \multirow{3}{*}{{100\%}}
& {Vanilla} & {\pmstd{82.09}{3.97}} & {\pmstd{69.81}{5.56}} & {\pmstd{4.18}{0.71}} & {\pmstd{1.22}{0.27}} \\
& & {PEFT   } & {\pmstd{81.81}{3.19}} & {\pmstd{69.34}{4.52}} & {\pmstd{4.50}{1.00}} & {\pmstd{1.36}{0.37}} \\
& & {LoRA   } & {\pmstd{81.10}{3.96}} & {\pmstd{68.40}{5.52}} & {\pmstd{5.02}{2.40}} & {\pmstd{1.50}{0.76}} \\
\midrule
\multirow{12}{*}{ImageTBAD} 
& {0\%} & {SAM-Med3D} & {55.85\tiny{$\pm$21.20}} & {42.41\tiny{$\pm$18.96}} & {14.19\tiny{$\pm$11.74}} & {3.65\tiny{$\pm$4.56}} \\
\cmidrule(lr){2-7}
& \multirow{4}{*}{10\%}
& Vanilla & \underline{\pmstd{41.57}{1.53}} & \underline{\pmstd{27.61}{1.12}} & \underline{\pmstd{16.79}{0.74}} & \underline{\pmstd{3.91}{0.37}} \\
& & PEFT    & \pmstd{40.14}{0.95} & \pmstd{26.51}{0.64} & \pmstd{18.36}{0.80} & \pmstd{4.23}{0.22} \\
& & LoRA    & \pmstd{35.94}{1.62} & \pmstd{23.05}{1.36} & \pmstd{20.07}{1.06} & \pmstd{5.99}{0.65} \\
& & \cellcolor{lightblue!50}\textbf{UnCoL (Ours)} & \cellcolor{lightblue!50}\textbf{\pmstd{70.97}{12.58}} & \cellcolor{lightblue!50}\textbf{\pmstd{57.17}{13.45}} & \cellcolor{lightblue!50}\textbf{\pmstd{4.83}{3.16}} & \cellcolor{lightblue!50}\textbf{\pmstd{1.30}{0.90}} \\
\cmidrule(lr){2-7}
& \multirow{4}{*}{20\%}
& Vanilla & \underline{\pmstd{43.93}{1.24}} & \underline{\pmstd{29.56}{0.88}} & \underline{\pmstd{15.34}{1.63}} & \pmstd{3.57}{0.26} \\
& & PEFT    & \pmstd{43.46}{1.07} & \pmstd{29.13}{0.78} & \pmstd{17.53}{1.22} & \underline{\pmstd{3.56}{0.34}} \\
& & LoRA    & \pmstd{42.81}{2.34} & \pmstd{28.48}{1.84} & \pmstd{19.17}{2.01} & \pmstd{3.89}{0.52} \\
& & \cellcolor{lightblue!50}\textbf{UnCoL (Ours)} & \cellcolor{lightblue!50}\textbf{\pmstd{78.89}{14.24}} & \cellcolor{lightblue!50}\textbf{\pmstd{67.13}{17.16}} & \cellcolor{lightblue!50}\textbf{\pmstd{3.70}{2.70}} & \cellcolor{lightblue!50}\textbf{\pmstd{1.01}{0.86}} \\
\cmidrule(lr){2-7}
& \multirow{3}{*}{{100\%}}
& {Vanilla} & {\pmstd{46.27}{14.04}} & {\pmstd{31.20}{12.17}} & {\pmstd{14.39}{8.55}} & {\pmstd{3.28}{1.33}} \\
& & {PEFT   } & {\pmstd{47.81}{1.25}} & {\pmstd{32.64}{0.87}} & {\pmstd{13.93}{1.17}} & {\pmstd{3.07}{0.10}} \\
& & {LoRA   } & {\pmstd{45.55}{1.48}} & {\pmstd{30.78}{1.22}} & {\pmstd{15.28}{1.37}} & {\pmstd{3.02}{0.34}} \\
\bottomrule
\end{tabular}}
\label{tab:FT}
\end{table}

Experiments in Table~\ref{tab:FT} follow the fine-tuning paradigm summarized in~\cite{gu2024build}, which provides a unified analysis of classical adaptation strategies for medical foundation models.
Our method demonstrates two practical advantages over these tuning baselines.
First, {
as a semi-supervised approach, UnCoL exhibits superior label efficiency. Under the same semi-supervised protocol, it surpasses fine-tuning baselines on both OASIS and ImageTBAD, even when those baselines are trained with partial or full annotations, highlighting its robustness to complex anatomy and severe distribution shifts.
On Pancreas-CT, UnCoL achieves clear gains over fine-tuning methods in partially labeled regimes, particularly on region-overlap metrics (Dice and Jaccard), while the boundary-based metrics (95HD and ASD) are comparable to LoRA under partial and full supervision, indicating similar boundary localization accuracy.}
Second, our method operates in a fully prompt-free manner, whereas the SAM-based fine-tuning baselines require test-time prompts (e.g., bounding boxes or points), introducing additional annotation overhead.
{Taken together, UnCoL combines strong label efficiency with prompt-free deployment, offering a practical alternative to fine-tuning-based adaptation in semi-supervised medical image segmentation.}

\subsection{Uncertainty Evaluation and Calibration Analysis}

\begin{figure}[htbp]
    \centering
    \begin{subfigure}[htbp]{\linewidth}
        \centering
        \includegraphics[width=\linewidth]{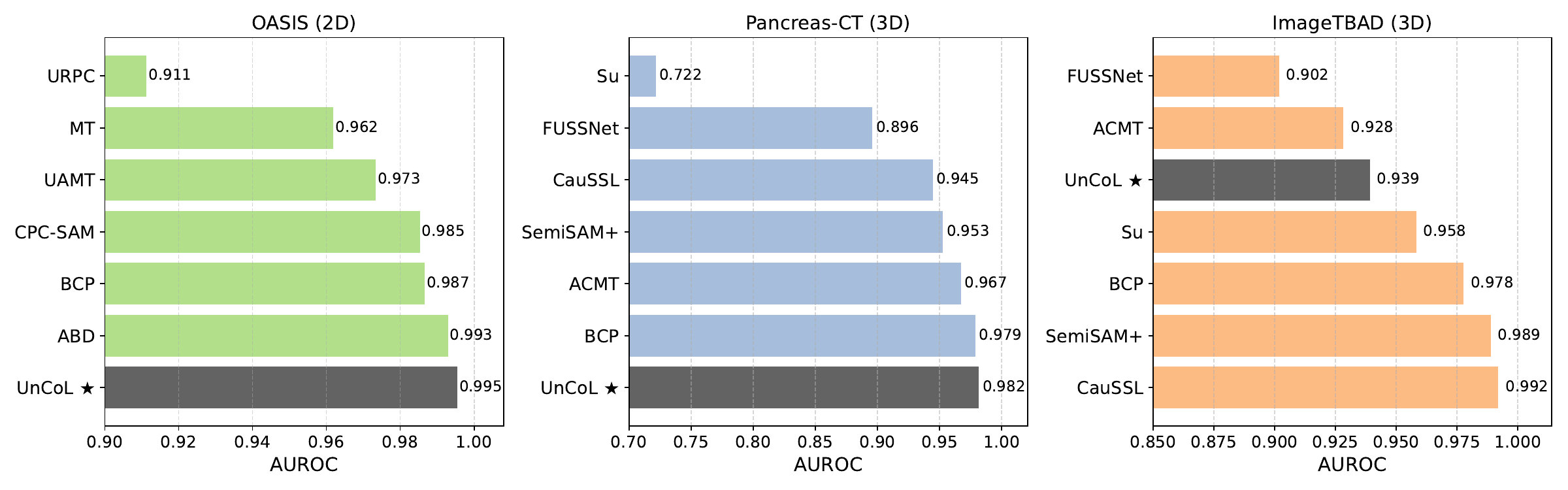}
        \caption{AUROC comparison based on entropy-ranked uncertainty.}
        \label{fig:auroc}
    \end{subfigure}

    \vspace{2pt}

    \begin{subfigure}[htbp]{\linewidth}
        \centering
        \includegraphics[width=\linewidth]{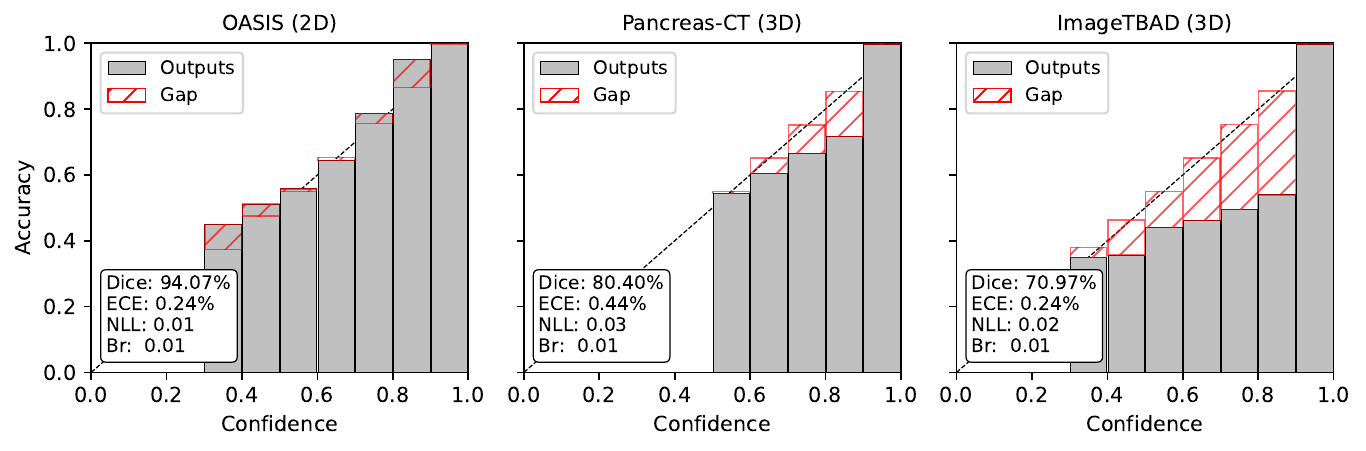}
        \caption{{Reliability diagrams.}}
        \label{fig:reliability}
    \end{subfigure}
    
    \vspace{2pt}

    \begin{subfigure}{\linewidth}
        \centering
        \includegraphics[width=\linewidth]{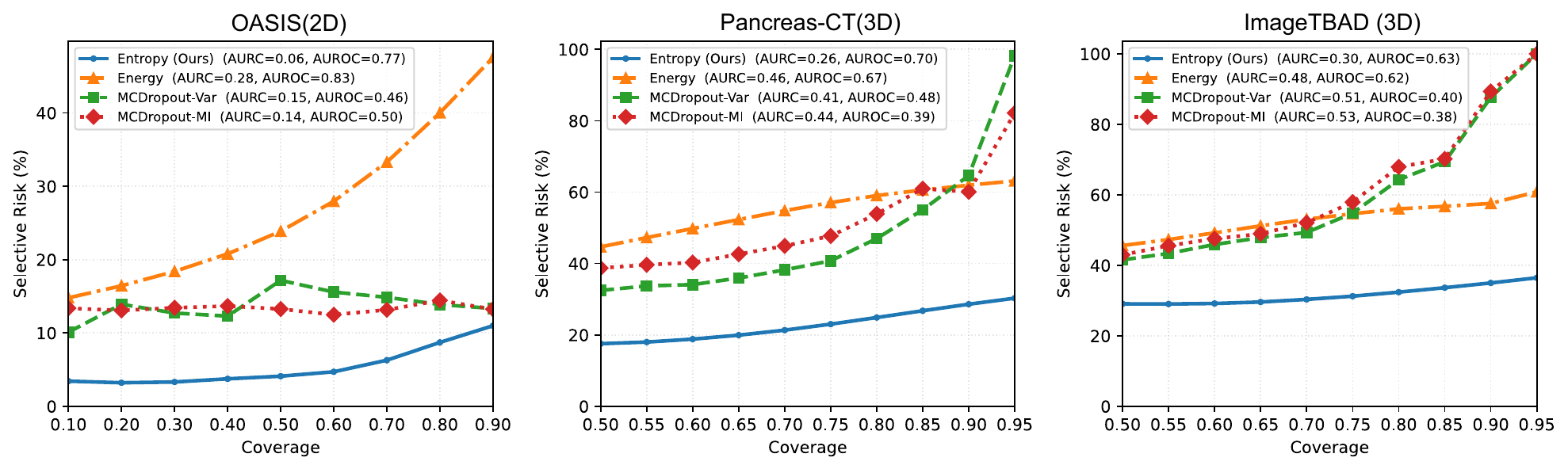}
        \caption{{Risk-coverage curves.}}
        \label{fig:riskcov}
    \end{subfigure}
    
    \caption{
Uncertainty evaluation on three datasets with 10\% labeled data. 
(a) AUROC comparison based on entropy-ranked uncertainty. 
{(b) Reliability diagrams (10 bins) showing accuracy versus confidence, with gray bars for bin-wise accuracy and red hatched gaps from perfect calibration. Lower ECE/NLL/Brier indicates better calibration.
(c) Risk-coverage curves comparing four uncertainty metrics: Shannon Entropy (Ours), Energy, MC Dropout-based Variance, and Mutual Information. Lower AURC and higher AUROC indicate better alignment between uncertainty and prediction error. }
}
    \label{fig:uncertainty_eval}
\end{figure}

{We evaluate the quality of UnCoL’s uncertainty estimates through three complementary perspectives: \textit{discriminability}, \textit{calibration}, and \textit{selective prediction reliability}, as illustrated in Fig.~\ref{fig:uncertainty_eval}. These analyses quantify how well uncertainty values reflect voxel-wise prediction errors, correspond to true confidence levels, and support reliable pseudo-label selection.}

\subsubsection{{Discriminability}}
{The discriminative power of uncertainty is measured using the area under the ROC curve (AUROC), which quantifies how well uncertainty distinguishes between correct and incorrect predictions. Higher AUROC indicates stronger correlation between uncertainty and voxel-wise error.}
As shown in Fig.~\ref{fig:auroc}, UnCoL achieves the {highest AUROC} on OASIS (0.995) and Pancreas-CT (0.982), and performs competitively on ImageTBAD (0.939). 
{The relatively lower AUROC on ImageTBAD is consistent with domain shifts and complex anatomical variations. Nonetheless, UnCoL still achieves the best Dice and boundary metrics on this dataset (Table~\ref{tab:tbad}), indicating that uncertainty estimation remains reliable under challenging conditions. Overall, these results show that our dual-teacher framework improves both segmentation accuracy and uncertainty discriminability.}

\subsubsection{{Calibration}}
{Reliability diagrams visualize the relationship between predictive confidence and observed accuracy~\cite{guo2017calibration}. Perfect calibration corresponds to points lying on the diagonal (confidence = accuracy).
As shown in Fig.~\ref{fig:reliability}, UnCoL exhibits strong calibration across all datasets, with small confidence deviations and low calibration errors measured by Expected Calibration Error (ECE $<0.5\%$), Negative Log-Likelihood (NLL $<0.03$), and Brier score ($<0.02$).
These quantitative measures confirm that UnCoL’s predicted confidence values align well with voxel-level correctness.}

\subsubsection{{Selective prediction reliability}}
{Risk-coverage curves evaluate how uncertainty supports selective prediction by progressively excluding high-uncertainty voxels~\cite{corbiere2019addressing}. A reliable uncertainty measure yields lower selective risk at any target coverage.
As shown in Fig.~\ref{fig:riskcov}, UnCoL’s entropy-based uncertainty consistently achieves the lowest risk compared with energy scores, MC Dropout variance, and MC Dropout-based mutual information, indicating that low-uncertainty regions tend to correspond to more reliable predictions.
}{Collectively, these results indicate that UnCoL produces uncertainty estimates that are both \textit{discriminative} and \textit{well-calibrated}, effectively supporting pseudo-label selection and consistency regularization under domain-shifted and data-scarce scenarios.}

\subsection{Computational Efficiency and Resource Consumption} \label{sec:efficiency}
\begin{table}[htbp]
  \centering
  \caption{Efficiency comparison of representative 2D/3D models.
  {We report parameter count, multiply-accumulate operations (MACs), peak GPU memory, average training time per iteration, and inference time per sample.
  UNet-based (UNet, MT, UAMT, BCP) and VNet-based (VNet, MT, UAMT, AC-MT, BCP, SemiSAM+) methods share identical backbones and comparable computational settings within each group.}
  }
  \label{tab:compute-profile}
  \resizebox{\linewidth}{!}{
  \begin{tabular}{c l
  S[table-format=3.2]
  S[table-format=6.2]
  S[table-format=6.2]
  r
  r}
    \toprule[1.3pt]
    \multirow{2}{*}{Dimension} & \multirow{2}{*}{Method} & \multicolumn{1}{c}{Params} & \multicolumn{1}{c}{MACs} & \multicolumn{1}{c}{Memory}  & \multicolumn{1}{c}{\scriptsize{Training Time}} & \multicolumn{1}{c}{\scriptsize{Inference Time}}  \\
     &  & \multicolumn{1}{c}{(M)} & \multicolumn{1}{c}{(G)} & \multicolumn{1}{c}{(MB)}  & \multicolumn{1}{c}{\scriptsize{(s/iteration)}} & \multicolumn{1}{c}{\scriptsize{(ms/sample)}}  \\
    \midrule
    \multirow{8}{*}{2D}
      & SAM~\cite{kirillov2023segment}~/~MedSAM~\cite{ma2024segment}        & 93.74 & 371.46 & 2771.71 & - & 102.20  \\ \cmidrule(lr){2-7}
      & UNet~\cite{ronneberger2015u} Group  &  1.81 &   3.01 &  145.96 &  0.16-0.51 & 1.86 \\
      & nnU-Net~\cite{isensee2021nnu}            & 18.69 &   4.76 &  211.52 & 0.36 &   4.16  \\
      & URPC~\cite{luo2022semi}              &  1.82 &   3.05 &  151.30 & 0.14 &  2.07 \\
      & ABD~\cite{chi2024adaptive}               & 27.15 &   7.75 &  162.94 & 0.54 & 13.71 \\ \cmidrule(lr){2-7}
      & CPC-SAM~\cite{miao2024cross}           & 8.56 &  32.78 &  412.60 & 0.23 &  17.59 \\
      & \cellcolor{lightblue!50}{UnCoL (Ours)}  & \multicolumn{1}{r}{\cellcolor{lightblue!50}{3.53}}  & \multicolumn{1}{r}{\cellcolor{lightblue!50}{3.60}}   & \multicolumn{1}{r}{\cellcolor{lightblue!50}{163.00}} & \cellcolor{lightblue!50}{0.81}  & \multicolumn{1}{r}{\cellcolor{lightblue!50}{4.34}} \\
    \midrule 
    \multirow{10}{*}{3D}
      & SAM-Med3D~\cite{wang2023sam}         & 100.51 &  186.44 & 1477.58 & - &  65.30 \\ \cmidrule(lr){2-7}
      & VNet~\cite{milletari2016v} Group &   9.44 &  97.91 &  810.74 & 0.45-1.54 & 14.93 \\
      & nnU-Net~\cite{isensee2021nnu}            &  30.45 & 215.48 & 1134.37 & 1.13 & 35.68 \\
      & Swin-UNETR~\cite{hatamizadeh2021swin}        &  61.99 & 785.06 & 3980.14 & 0.41 & 194.09 \\
      & MC-Net+~\cite{wu2022mutual}           &  15.25 & 298.96 & 1120.84 & 2.52 &  57.79 \\
      & CauSSL~\cite{miao2023caussl}            &  12.98 & 190.70 & 1576.71 & 0.98 & 43.13 \\
      & Su \emph{et al.}~\cite{su2024mutual}  &  12.35 & 198.83 & 1221.75  & 0.33 &  39.28 \\ \cmidrule(lr){2-7}
      & \cellcolor{lightblue!50}{UnCoL (Ours)} & \multicolumn{1}{r}{\cellcolor{lightblue!50}{13.94}} & \multicolumn{1}{r}{\cellcolor{lightblue!50}{144.45}} & \multicolumn{1}{r}{\cellcolor{lightblue!50}{887.97}}  &  \cellcolor{lightblue!50}{1.10} & \multicolumn{1}{r}{\cellcolor{lightblue!50}{45.87}} \\
    \bottomrule[1.3pt]
  \end{tabular}}
\end{table}

{Table~\ref{tab:compute-profile} compares the computational efficiency of different dimensional settings (2D and 3D) and learning paradigms{, reporting parameter count, MACs, peak GPU memory, average training time per iteration (default batch size), and inference time per sample. Under the 2D setting, training time is measured on the OASIS dataset with 5\% labeled data, and under the 3D setting on the Pancreas-CT dataset with 10\% labeled data. }
For fair comparison, methods sharing 
{identical backbones and comparable computational configurations are grouped together (e.g., MT/UAMT/BCP under the U-Net or V-Net backbone), and the reported training time reflects the range observed within each group.}
Compared with lightweight baselines, UnCoL incurs only moderate computational overhead while achieving substantially higher accuracy (see Sec.~\ref{sec:quantresults}), 
{demonstrating its practical effectiveness under constrained computational budgets.}
Moreover, it requires significantly less memory and {inference time} than zero-shot or fine-tuned foundation models such as CPC-SAM, highlighting its practicality for efficient semi-supervised segmentation.}

\section{Discussion and Conclusion}
\label{sec:dis}
This work presents UnCoL, an uncertainty-informed collaborative learning framework designed to harmonize generalization and specialization for semi-supervised medical image segmentation. By integrating a frozen foundation model as a generalized teacher and a dynamically evolving specialized teacher, UnCoL enables prompt-free adaptation via dual-path knowledge distillation and uncertainty-aware pseudo-labeling. 
{Extensive experiments across three datasets and varying label ratios show that UnCoL consistently outperforms zero-shot foundation models, conventional semi-supervised methods, and even fully supervised fine-tuning in certain cases.
Ablation studies confirm that the dual-teacher design and uncertainty-aware supervision jointly enhance learning stability, pseudo-label reliability, and representation consistency, enabling the student model to effectively integrate complementary knowledge while avoiding supervision collapse.
}

{Overall, UnCoL provides an annotation-efficient and computationally lightweight solution for medical image segmentation, offering practical value in clinical scenarios where labeled data are scarce and adaptation cost must be minimized. Future work will extend this framework to multimodal and cross-domain settings, address class imbalance, and explore explicit data-level uncertainty modeling to further broaden its applicability in real-world medical imaging workflows.}

\bibliographystyle{ieeetr}
\bibliography{mybibliography}

\end{document}